\documentclass[lettersize,journal,monochrome]{IEEEtran}

\usepackage{amsmath,amsfonts, amssymb}
\usepackage{mathtools}

\usepackage{array}
\usepackage[caption=false,font=normalsize,labelfont=sf,textfont=sf]{subfig}
\usepackage{textcomp}
\usepackage{stfloats}
\usepackage{url}
\usepackage{verbatim}
\usepackage{graphicx}
\usepackage{cite}

\usepackage{tikz}
\usepackage{comment}
\usepackage{epsfig}
\usepackage{booktabs} 
\usepackage{soul}
\usepackage{capt-of}
\usepackage{multirow}
\usepackage{tablefootnote}

\usepackage{algorithm}
\usepackage{algpseudocode}

\usepackage{fancyhdr}
\usepackage{setspace}
\usepackage{dsfont}
\usepackage{pifont}
\newcommand{\cmark}{\ding{51}}%
\newcommand{\xmark}{\ding{55}}%
\usepackage[pagebackref=true,breaklinks=true,letterpaper=true,colorlinks=false,bookmarks=false]{hyperref}
\hypersetup{
    urlcolor=magenta,
}
\begin{document}

\title{Adversarial Image Color Transformations in Explicit Color Filter Space}

\author{Zhengyu~Zhao,
        Zhuoran~Liu,
        and~Martha~Larson
\thanks{Zhengyu Zhao is with Xi'an Jiaotong University, Xi'an, China, and CISPA Helmholtz Center for Information Security, Saarbrücken, Germany.}
\thanks{Zhuoran Liu is with the Institute for Computing and Information Sciences, Radboud University, Nijmegen, The Netherlands.}
\thanks{Martha Larson is with the Institute for Computing and Information Sciences, Radboud University, Nijmegen, The Netherlands.}
}



\maketitle

\begin{abstract}

Deep Neural Networks have been shown to be vulnerable to adversarial images.
Conventional attacks strive for indistinguishable adversarial images with strictly restricted perturbations.
Recently, researchers have moved to explore distinguishable yet non-suspicious adversarial images and demonstrated that color transformation attacks are effective.
In this work, we propose Adversarial Color Filter (AdvCF), a novel color transformation attack that is optimized with gradient information in the parameter space of a simple color filter.
In particular, our color filter space is explicitly specified so that we are able to provide a systematic analysis of model robustness against adversarial color transformations, from both the attack and defense perspectives.
In contrast, existing color transformation attacks do not offer the opportunity for systematic analysis due to the lack of such an explicit space.
We further demonstrate the effectiveness of our AdvCF in fooling image classifiers and also compare it with other color transformation attacks regarding their robustness to defenses and image acceptability through an extensive user study.
We also highlight the human-interpretability of AdvCF and show its superiority over the state-of-the-art human-interpretable color transformation attack on both image acceptability and efficiency.
Additional results provide interesting new insights into model robustness against AdvCF in another three visual tasks.

\end{abstract}


\maketitle
\IEEEdisplaynontitleabstractindextext

\IEEEpeerreviewmaketitle

\begin{figure*}[t]
\begin{center}
  \includegraphics[width=0.9\textwidth]{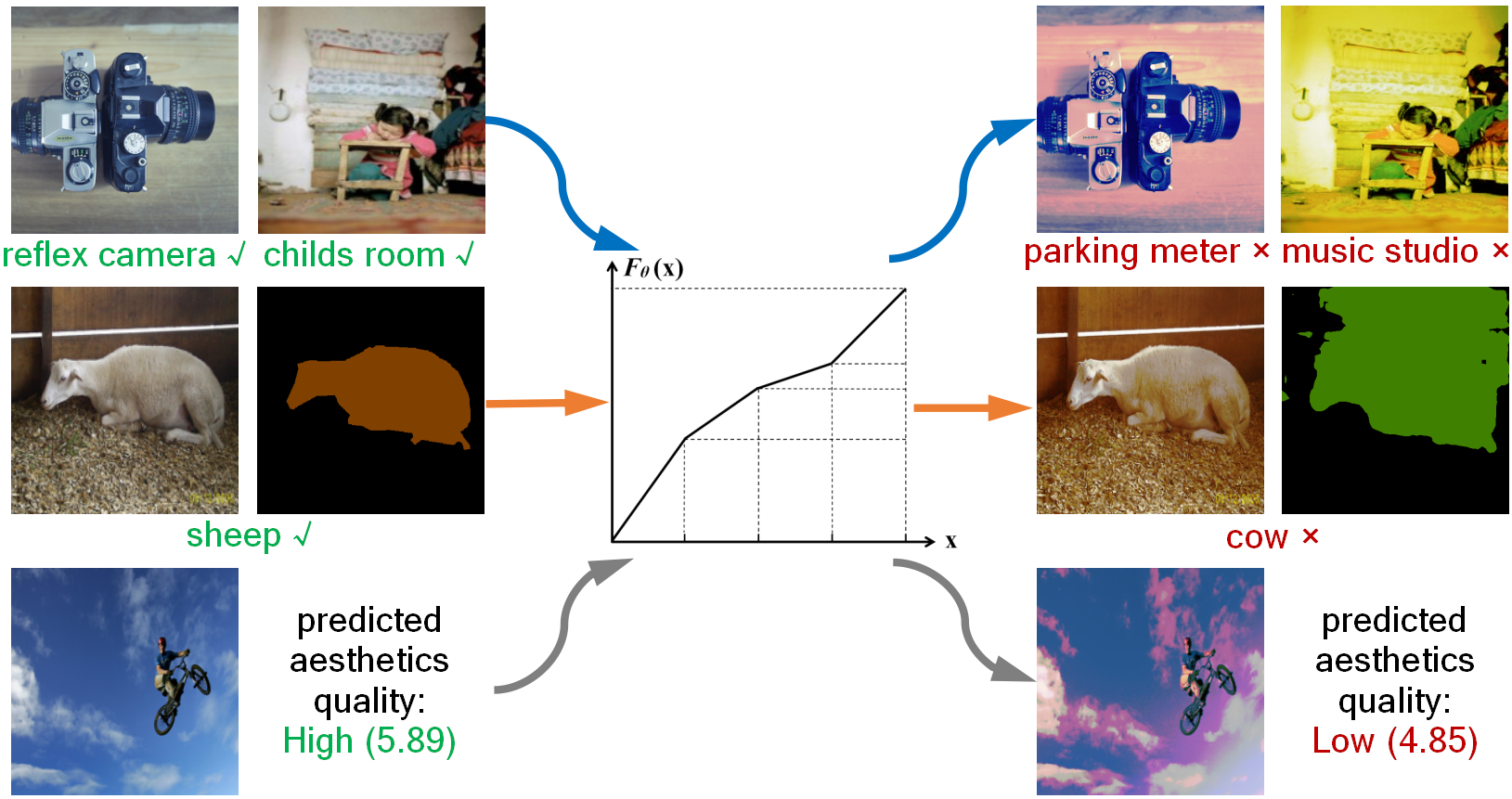}
\end{center}
\caption{Overview of our Adversarial Color Filter (AdvCF), which makes it possible to generate non-suspicious adversarial images in different computer vision tasks: from top to bottom, object/scene classification (top row), semantic segmentation (middle row), and aesthetic quality assessment (bottom row).
Additional image examples can be found in Appendix A.
}
\label{fig:overview}
\end{figure*}

\section{Introduction}\label{sec:introduction}

\IEEEPARstart{D}{espite} the great success of Deep Neural Network (DNN)-powered computer vision models, recent research has shown that they are remarkably vulnerable to adversarial images~\cite{szegedy2013intriguing}, i.e., intentionally modified images that cause incorrect model predictions.
Such vulnerability has been extensively studied in basic image classification~\cite{carlini2017towards,goodfellow2014explaining,papernot2016limitations}, but also other tasks, such as object detection~\cite{xie2017adversarial,zhao2019seeing}, semantic segmentation~\cite{arnab2018robustness,xie2017adversarial}, and retrieval~\cite{liu2019s,tolias2019targeted}.

Related work has commonly assumed that adversarial images must be stealthy~\cite{gilmer2018motivating}.
Most adversarial attacks strive for stealthiness by following the assumption that the adversarial image should be \emph{indistinguishable} from its corresponding original image~\cite{szegedy2013intriguing,goodfellow2014explaining}.
More recently, this ``indistinguishable'' assumption has been questioned because in real-world scenarios adversarial images are encountered in isolation, i.e., they are not examined with the corresponding original images as reference~\cite{gilmer2018motivating}.
For this reason, increasingly more recent attacks have been aimed to achieve distinguishable yet \emph{non-suspicious} perturbations, which are allowed to be large but do not draw human attention because they modify groups of pixels along dimensions consistent with human interpretation of images.
Among these non-suspicious attacks, approaches~\cite{bhattad2020Unrestricted,hosseini2018semantic,Laidlaw2019functional,shamsabadi2020colorfool} that rely on color transformations have shown more promising results in general scenarios compared to those based on limited, domain-specific attributes~\cite{eykholt2017robust,joshi2019semantic,qiu2019semanticadv,sharif2016accessorize}.

In this work, following this promising trend, we propose a new color transformation attack, called Adversarial Color Filter (AdvCF), which models the color transformation space as the parameter space of a simple color filter~\cite{hu2018exposure}.
Specifically, adversarial images are generated by optimizing the $L_{\infty}$-bounded filter parameters using gradient information.
The novelty of our AdvCF is twofold.

Firstly, AdvCF provides an \emph{explicit} color filter space, in which the filter parameters can be explicitly adjusted during optimization.
This makes it possible to conduct a systematic robustness analysis of image classifiers against adversarial color transformations from both the attack and defense perspectives~\cite{alaifari2018adef,laidlaw2020perceptual}.
This possibility is useful because a systematic robustness analysis is a key to understanding the model vulnerability to specific adversarial attacks~\cite{croce2021robustbench,wu2022blackboxbench}.
Without a systematic robustness analysis, the attacker's potential might not be fully uncovered, and as a result the model robustness might be overestimated~\cite{athalye2018obfuscated,carlini2019evaluating,tramer2020adaptive}.
Different from our AdvCF, the current state-of-the-art color transformation attacks, ReColorAdv~\cite{Laidlaw2019functional} and cAdv~\cite{bhattad2020Unrestricted}, do not offer the possibility for a systematic robustness analysis mainly because they parameterize their color transformations in one space but constrain them in a different space.
ReColorAdv is parameterized based on discrete color grids, but constrained using implicit regularization on image smoothness and color transformation-agnostic, pixel-wise $L_p$ bounds. 
cAdv is parameterized based on the colorization process of a DNN model, which is non-transparent, and does not imposed any bound constraints, but rather is controlled by varying the number and locations of sparse input color hints.

Secondly, AdvCF is directly interpretable to humans because its implementation mimics the color curve adjustment in the common photo retouching process, and can be achieved with just standard photo editing software.
Studying human-interpretable attacks is necessary since the non-suspiciousness of adversarial images is defined with respect to human judgment.
It is also important in practical use scenarios, e.g., using adversarial images for protecting users' privacy~\cite{cherepanova2021lowkey,liu2019s,ppoverview2018,rajabi2021practicality}.
Fig.~\ref{fig:overview} provides an overview of our AdvCF in different tasks.
Different from our AdvCF, ReColorAdv~\cite{Laidlaw2019functional} and cAdv~\cite{bhattad2020Unrestricted} are not human-interpretable due to the use of additional, sophisticated, operations (e.g., the color interpolation and smoothness regularization)~\cite{Laidlaw2019functional} or the non-transparent colorization process of a DNN model~\cite{bhattad2020Unrestricted}.
Moreover, the current state-of-the-art human-interpretable attack, ColorFool\cite{shamsabadi2020colorfool}, results in adversarial images with low acceptability (see Section~\ref{sec:compare} for more details), and it is not efficient due to the use of random search.

In sum, our work makes the following contributions:
\begin{itemize}
  \setlength{\parskip}{0pt}
  \setlength{\itemsep}{0pt}
\item We propose Adversarial Color Filter (AdvCF), a new color transformation attack based on optimizing a simple color filter.
Benefiting from its explicit color transformation space, we conduct a systematic robustness analysis of image classifiers against AdvCF from both the attack and defense perspectives.

\item We demonstrate the effectiveness of our AdvCF in fooling image classifiers and then comprehensively compare it with other color transformation attacks on their robustness to defenses and image acceptability through a user study.
We also further improve AdvCF in terms of its image acceptability by using color style guidance. 

\item We demonstrate that our AdvCF improves upon the current state-of-the-art human-interpretable color transformation attack (i.e., ColorFool~\cite{shamsabadi2020colorfool}) on both the image acceptability and efficiency.

\item We provide new insights into understanding model robustness against AdvCF in other tasks, including scene recognition, semantic segmentation, and previously unexplored, aesthetic quality assessment.
\end{itemize}

This paper extends our previous conference paper~\cite{zhao2020adversarial} mainly by 1) re-designing the original attack to have an explicit color transformation space, making our new systematic robustness analysis possible, 2) adding analysis also from the defense perspective, 3) adding a user study on the image acceptability of different color transformation attacks, and 4) adding experiments in another three tasks.
Code is publicly available at~\url{https://github.com/ZhengyuZhao/AdvCF/tree/master/Journal_version}.

\section{Related Work}
In this section, we discuss related work on adversarial attack and defense techniques.
For precision, we first provide a formal definition of the creation of adversarial images using the basic image classification as an example.
A neural network can be denoted as a function $\mathcal{F}(\boldsymbol{x})=l$ that takes an image $\boldsymbol{x}\in\mathbb{R}^n$ as input and predicts an original label $l$ for it.
An attack aims to induce a misclassification by modifying the original image $\boldsymbol{x}$ into $\boldsymbol{x}'$.
Accordingly, the attack is regarded as successful should $\boldsymbol{x}'$ be assigned any new class label other than the original one, i.e., $\mathcal{F}(\boldsymbol{x}')\neq{l}$.

\subsection{Adversarial Attacks}
\label{sec:attacks}
In this subsection, we review adversarial attacks under both the indistinguishable and non-suspicious assumptions.

\subsubsection{Attacks with Indistinguishable Adversarial Images}
Most adversarial attacks against DNN models are based on the assumption that the created adversarial image should be indistinguishable from the corresponding original image.
The earliest work~\cite{szegedy2013intriguing} optimizes the joint objective of the misclassification, which is based on the cross-entropy loss, and the $L_{2}$ distance.
C\&W~\cite{carlini2017towards} followed a similar formulation but relies on a more effective, logit loss.
Additionally, the box constraint was eliminated by introducing a new variable.
The attack optimization can be expressed as:
\begin{equation}
\label{cw}
\underset{\boldsymbol{w}}{\mathrm{min}}
~~{\|\boldsymbol{x}'-\boldsymbol{x}\|}_2^2+\lambda L_{\mathrm{CW}}(\boldsymbol{x}',l),
\end{equation}
where
\begin{equation}
\label{cw2}
\begin{gathered}
 L_{\mathrm{CW}}(\boldsymbol{x}')=\mathrm{max}(Z(\boldsymbol{x}')_l-{\mathrm{max}}\{Z(\boldsymbol{x}')_i:i\neq l\},-\kappa),\\
\textrm{and}~~\boldsymbol{x}'=\frac{1}{2}(\tanh(
\mathrm{arctanh}(\boldsymbol{x})+\boldsymbol{w})+1).
\end{gathered}
\end{equation}
Here, $L_{\mathrm{CW}}$ is the adversarial loss, $\boldsymbol{w}$ is the new variable, and $Z(x')_i$ is the logit of the $i$-th class for the modified image $\boldsymbol{x}'$.
The confidence level of the misclassification that needs to be reached can be varied by adjusting $\kappa$.

Due to the necessity of line search for an optimal $\lambda$, such joint optimization suffers from high computational cost.
To address this issue, other studies~\cite{goodfellow2014explaining,kurakin2016adversarial,madry2017towards,rony2019decoupling} have built up their approaches based on Projected Gradient Descent (PGD), where the perturbations are projected into a small pre-defined $L_p$-norm ball around the original image to ensure indistinguishability, i.e., satisfying $\|\boldsymbol{x}'-\boldsymbol{x}\|_{p} \leq \epsilon$.
Specifically, early work~\cite{goodfellow2014explaining} proposed the Fast Gradient Sign Method (FGSM), which takes a single step of gradient descent.
FGSM was directly extended to an iterative variant, I-FGSM~\cite{kurakin2016adversarial,madry2017towards}, 
which is stronger because it exploits finer gradient information.
I-FGSM can be formulated as:
\begin{equation}
\label{IFGSM}
{\boldsymbol{x}_0'}=\boldsymbol{x},~~{\boldsymbol{x}_{t+1}'}={\boldsymbol{x}_{t}'}+\alpha\cdot\ \mathrm{sign}({\nabla_{{\boldsymbol{x}}}J(\boldsymbol{x}_{t}',l)}),
\end{equation}
where $\boldsymbol{x}_{t}'$ is the perturbed image at the $t$-th iteration updated with the step size $\alpha$.
The cross-entropy loss is widely used for the adversarial loss $J(\cdot,\cdot)$.
The $L_{\infty}$ norm is commonly used and other norms have also been explored~\cite{rony2019decoupling,Croce_2019_ICCV}.

Since $L_p$ norms are known to be inadequate in measuring indistinguishability, there have been some recent studies relying on more human-vision-aligned measurements.
Most studies in this direction have exploited textural information in images.
Specifically, more advanced similarity metrics, such as the well-known Structural SIMilarity (SSIM) metric~\cite{wang2004image} was explored in~\cite{rozsa2016adversarial}.
Other methods~\cite{Croce_2019_ICCV,luo2018towards,zhang2020smooth} adapt the $L_p$ norms to local textural patterns, i.e., hiding perturbations in image regions that have high visual variation.
Non-additive perturbations, such as spatial transformations~\cite{xiao2018spatially,kanbak2018geometric,alaifari2018adef,engstrom2019exploring,wong2019wasserstein} have also been explored.
Beyond textural information, perceptual color distance can also be used to improve indistinguishability~\cite{zhao2020towards}.

\subsubsection{Attacks with Non-Suspicious Adversarial Images}
\label{sec:non-sus}
The assumption that adversarial images should be indistinguishable has been argued to lack compelling motivation in real-world scenarios~\cite{gilmer2018motivating}.
When there is no direct comparison to the original images, adversarial images can remain non-suspicious without strictly restricting the perturbations.
Following this new assumption, recent studies have explored modifications of domain-specific attributes, such as adding adversarial eyeglass frame~\cite{sharif2016accessorize} or manipulating make-ups against face recognition~\cite{joshi2019semantic,qiu2019semanticadv}, and pasting adversarial stickers against road sign recognition~\cite{eykholt2017robust}.

Other studies have relied on modifying more general image attributes, e.g., image colors.
The earliest work on such color transformation attacks searches for possible adversarial images by randomly shifting the color components (Hue and Saturation) in the HSV color space~\cite{hosseini2018semantic}.
This attack process is human-interpretable because it does not involve any sophisticated operations.
However, the resulting adversarial images are not visually acceptable since the transformations are not constrained.

The later ColorFool~\cite{shamsabadi2020colorfool} follows a similar idea and has improved image acceptability by imposing additional constraints in image regions occupied by specific visual concepts (including person, sky, vegetation, and water) that are sensitive to color changes. 
These concepts are localized by a semantic segmentation model before attacking.
As a result, the image acceptability strongly depends on the segmentation quality and the specific visual concept categories.
In addition, using random color shifts requires a lot of trials to find successful adversarial images.

Differently, ReColorAdv~\cite{Laidlaw2019functional} and cAdv~\cite{bhattad2020Unrestricted} have proposed to parameterize the color transformation (in implicit attack spaces) and optimize it with gradient information, leading to more efficient attacks.
Although these two approaches sacrifice human-interpretability, they generate adversarial images more efficiently and make them more visually acceptable.
Specifically, ReColorAdv defines a functional color transformation, $f$, that operates over a color space $\mathcal{G}$ that is sampled for reasons of computational efficiency.
The optimization can be expressed as:
\begin{equation}
\label{recoloradv}
    \underset{f}{\mathrm{min}}~L_\textrm{CW}(f(\boldsymbol{x}),l) + \lambda L_\textrm{smooth}(f),
\end{equation}
where
\begin{equation}
\label{recons}
\begin{gathered}
L_\textrm{smooth}(f) = \sum_{g_j \in \mathcal{G}} \sum_{g_k \in \mathcal{N}(g_j)} \|(f(g_j) - g_j) - (f(g_k) - g_k)\|_2
 ,\\
\textrm{and}~~\|\boldsymbol{x}-f(\boldsymbol{x})\|_{\infty} \leq \epsilon.
\end{gathered}
\end{equation}
The additional regularization term $L_\textrm{smooth}$ accounts for local color smoothness by constraining the similarity between each point $g_j \in \mathcal{G}$ and $g_k$ in its neighborhood $\mathcal{N}$.
Specifically, trilinear interpolation is used to determine the values of the pixels that are not on the sampled color space.
Pixel-level $L_{\infty}$ bounds are used to restrict the color changes.

cAdv directly integrates the attack optimization into the process of automatically colorizing gray-scale images by using a DNN-based colorization model that has been pre-trained on massive image data~\cite{zhang2017real}.
The optimization can be expressed as:
\begin{equation}
\label{cadv}
\underset{M, X_{ab}}{\mathrm{min}}~ J(\mathcal{F}(\mathcal{C}(X_L, X_{ab}, M)), l),
\end{equation}
where $\mathcal{C}$ is the colorization model, and $X_L \in \mathbb{R} ^{H \times W \times 1}$ is the L channel of the image in CIELAB color space.
Specifically, the sparse input hints, $X_{ab} \in \mathbb{R} ^{H \times W \times 2}$, are taken from the original color image at locations determined by a binary mask $M \in \mathbb{B} ^{H \times W \times 1}$.
The number and locations of hints can be varied to implicitly control colorization.
In order to produce realistic colorization, cAdv is designed to preserve colors in image regions that have a relatively low tolerance for color changes.
To this end, a pixel-wise color distribution output by the colorization model is used to calculate the entropy of each pixel.
Any given image is segmented into 8 clusters based on the pixel values of the AB color channels, and then color hints will be sampled from low-entropy cluster(s) in order to guide the automatic colorization to preserve colors in image regions that fall in these clusters.
Note that cAdv only works for images with a size of $224\times224$ due to the fixed input resolution of the colorization model.

ReColorAdv does not provide an explicit color transformation space because its parameters cannot be explicitly adjusted.
Instead, it is implicitly controlled based on additional smoothness regularization and color transformation-agnostic, pixel-wise $L_p$ bounds.
In addition, the color interpolation and smoothness regularization in ReColorAdv make it not directly interpretable to humans.
cAdv also lacks an explicit color transformation space and human interpretability due to the use of a non-transparent neural network and no bound constraints.
Compared to ColorFool, the current state-of-the-art human-interpretable approach, our AdvCF can yield better results on both image acceptability (cf. Table~\ref{tab:user}) and efficiency (cf. Table~\ref{tab:trans}).

\subsection{Adversarial Defenses}
\label{sec:defenses}
As adversarial attacks raise increasing concerns, corresponding defenses have been extensively studied in both the black-box and white-box scenarios.
In the following, we review existing defenses against both indistinguishable and non-suspicious attacks.

\subsubsection{Defenses against Indistinguishable Attacks}
\label{sec:def_imp}
\noindent\textbf{Black-box Defenses.} Defenses against adversarial images have been mainly studied in the relatively easy, black-box scenario, where the attacker has no knowledge of the applied defenses.
Specifically, some defenses~\cite{grosse2017statistical,ma2018characterizing,metzen2017detecting} follow the conventional image forgery detection pipeline to train a binary image classifier to differentiate adversarial images from normal images.
Other defenses have proposed to filter out the adversarial perturbations using simple input transformations, such as JPEG compression~\cite{das2018shield}, feature squeezing~\cite{xu2017feature}, and random resizing\&padding~\cite{xie2017mitigating}.
These defenses do not require any model re-training and have little impact on the original model accuracy for clean images. 

\noindent\textbf{White-box Defenses.} However, in the more challenging, white-box scenario where the attack can be adapted to the specific defense black-box defenses were shown to provide little robustness~\cite{carlini2017adversarial,athalye2018obfuscated}.
Many recent studies have also proposed defenses that are specifically designed for white-box robustness but were also shown to be largely circumvented by simple adaptive attacks~\cite{tramer2020adaptive}.
This leaves adversarial training~\cite{madry2017towards,zhang2019theoretically} the only promising white-box defense approach.
Adversarial training involves model re-training with adversarial examples.
In the context of indistinguishable adversarial images, substantial model robustness has been achieved based on the iterative Projected Gradient Descent (PGD)~\cite{madry2017towards,zhang2019theoretically} or single-step FGSM with randomized initialization~\cite{wong2020fast}.
Adversarial training has also been explored for other types of indistinguishable adversarial images, e.g. spatial transformations~\cite{xiao2018spatially,alaifari2018adef,wong2019wasserstein}.
Recent work~\cite{laidlaw2021perceptual} has also achieved generalizable robustness against multiple types of indistinguishable adversarial images by using adversarial images that are bounded by the neural perceptual distance, LPIPS~\cite{zhang2018unreasonable}.

\subsubsection{Defenses against Non-Suspicious Attacks}
\label{def_non}
\noindent\textbf{Black-box Defenses.} Conventional input transformation-based methods are agnostic to the type of attack, and so they can be directly applied to defending against non-suspicious attacks.
Specifically, color transformation attacks have been shown to be more robust against commonly used input transformations than those under the indistinguishable assumption~\cite{shamsabadi2020colorfool,bhattad2020Unrestricted}.
\noindent\textbf{White-box Defenses.} For white-box robustness, recent work~\cite{wu2020defending} has demonstrated the effectiveness of adversarial
training in defending against the adversarial eyeglass frame~\cite{sharif2016accessorize} and adversarial stickers~\cite{eykholt2017robust}.
For defenses against adversarial color transformations, existing work has explored adversarial training~\cite{Laidlaw2019functional,shamsabadi2020colorfool} but not in an explicit color transformation space.

In this paper, for black-box defenses, we test diverse input transformation-based methods.
We especially consider an intuitively useful transformation, gray-scale conversion.
For white-box defenses, we explore adversarial training and discuss the relation between model robustness against color transformation attacks and the conventional $L_{\infty}$ robustness.

\subsection{Explicit Attack Space for Systematic Robustness Analysis}
\label{sec:exp}
Achieving a comprehensive understanding of adversarial robustness in various threat models requires conducting corresponding systematic robustness analyses~\cite{croce2021robustbench,wu2022blackboxbench}, and the key to conducting such systematic robustness analyses is an explicit attack space~\cite{alaifari2018adef,laidlaw2020perceptual}. 
In the research of indistinguishable adversarial images, explicit attack spaces have been mainly defined based on simple, $L_p$ bounds ~\cite{carlini2017towards,goodfellow2014explaining,kurakin2016adversarial,szegedy2013intriguing,papernot2016limitations}.
For other indistinguishable studies that involve more human-perception-aligned attacks~\cite{rozsa2016adversarial,luo2018towards,xiao2018spatially,kanbak2018geometric,alaifari2018adef,wong2019wasserstein,Croce_2019_ICCV,zhang2020smooth,zhao2020towards,laidlaw2021perceptual}, explicit attack spaces are based on the texture-aware pixel bound~\cite{Croce_2019_ICCV}, deforming vector field~\cite{alaifari2018adef}, Wasserstein distance~\cite{wong2019wasserstein}, or neural perceptual distance (LPIPS)~\cite{laidlaw2020perceptual}.

However, in the research of non-suspicious adversarial images~\cite{sharif2016accessorize,eykholt2017robust,hosseini2018semantic,joshi2019semantic,qiu2019semanticadv,Laidlaw2019functional,shamsabadi2020colorfool}, including those on adversarial color transformations~\cite{hosseini2018semantic,Laidlaw2019functional,shamsabadi2020colorfool}, there lack explicit attack spaces due to the arbitrary design of the attack approaches.
Differently, our work provides an explicit attack space that allows for conducting a systematic robustness analysis of adversarial color transformations.

\section{Adversarial Color Filter (AdvCF)}
In this section, we introduce Adversarial Color Filter (AdvCF), our new color transformation attack targeting non-suspicious adversarial images.
First, we briefly introduce the preliminary knowledge of parametric image filters and then present our new attack.

\subsection{Parametric Filters for Image Retouching}
Most state-of-the-art approaches~\cite{gharbi2017deep,isola2017image,zhu2017unpaired} to automatically retouch photos have mainly used DNNs to parameterize the editing process and are consequently not interpretable to users.
In order to make the retouching human-interpretable, recent work~\cite{hu2018exposure} has proposed to parameterize the process based on commonly used image filters.
Using image filters is also more efficient since they involve far fewer parameters and can be applied independently of the image resolution.

The color filter provided by~\cite{hu2018exposure} mimics the color curve adjustment that is widely applied in popular photo retouching software, such as Photoshop and Lightroom.
Its formulation can be expressed as:
\begin{equation}
\label{filter}
\begin{gathered}
F_{\boldsymbol{\theta}}(x_k)=\sum\limits_{i=1}^{k-1}\frac{\theta_i}{\theta_\mathrm{sum}}+(K \cdot x_k-(k-1)) \cdot \frac{\theta_k}{\theta_\mathrm{sum}},
\end{gathered}
\end{equation}
where $\theta_\mathrm{sum}=\sum\limits_{k=1}^{K}{\theta_k}$ and $K$ demotes the total number of pieces.
A given pixel $x_k$ whose value falls into the $k$-th piece of the function will be transformed using the parameter $\theta_k$, and $F_{\boldsymbol{\theta}}(x_k)$ is its corresponding output.

\begin{figure}[!t]
\begin{center}
  \includegraphics[width=0.7\columnwidth]{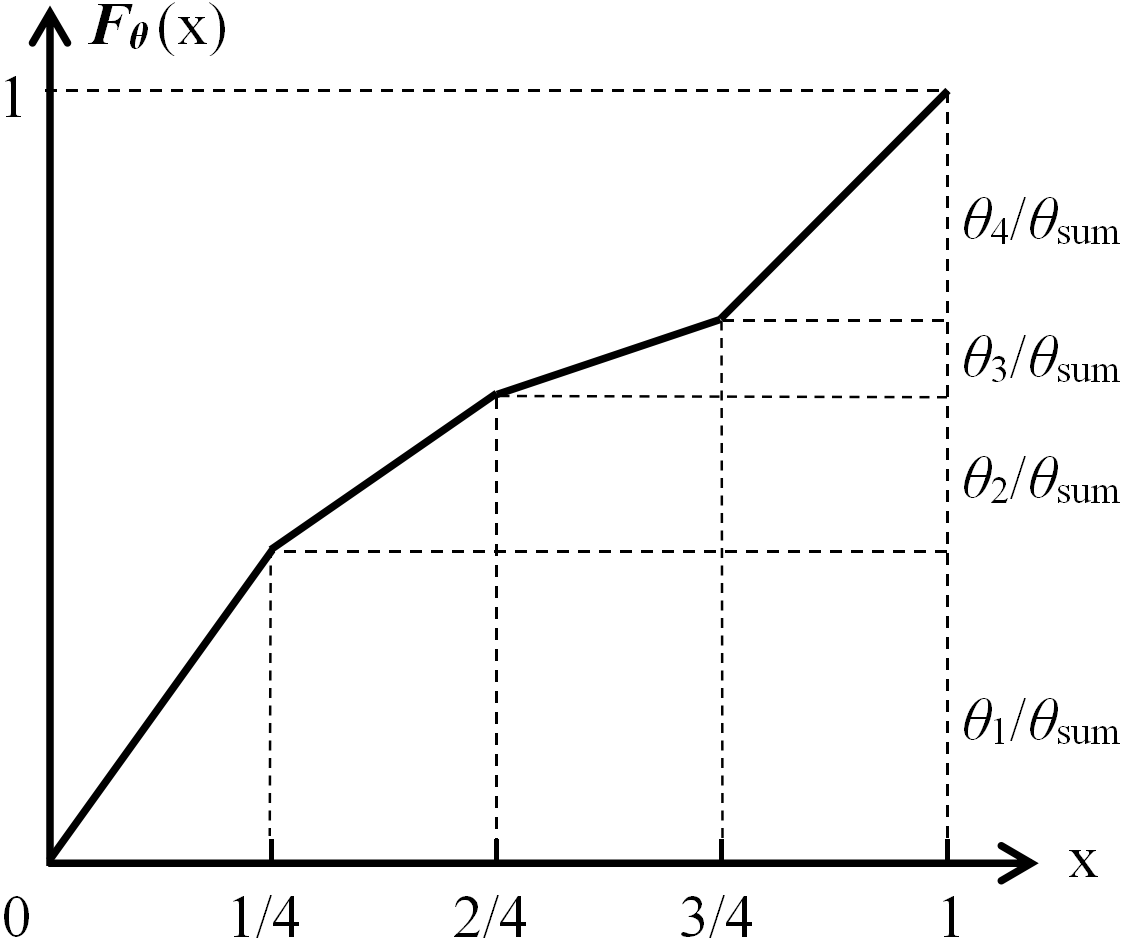}
\end{center}
   \caption{An illustration of the color filter used in our AdvCF (here, $K=4$ in Eq.~\ref{filter}).}
\label{fig:ill}
\end{figure}

By doing this, pixels with similar values will be transformed with the same parameter, which maintains color uniformity well.
Normally, the three RGB channels are processed independently.
Optimizing adversarial images within another color space that aligns better with human vision might further improve image acceptability~\cite{Laidlaw2019functional,zhao2020towards}, but we do not explore it in this work.
An example of this function with four pieces ($K=4$) is illustrated in Fig.~\ref{fig:ill}.
It is easy to see that the initial values of all $K$ parameters are equal to $1/K$.

\begin{algorithm}[!t]
\setstretch{1.0}
\caption{Adversarial Color Filter (AdvCF)}
\label{AdvCF}
\algrenewcommand\algorithmicrequire{\textbf{Input:}}
\algrenewcommand\algorithmicensure{\textbf{Output:}}
\algorithmicrequire{\\$\boldsymbol{x}$: original image, $l$: original label, $T$: iteration budget\\$F_{\boldsymbol{\theta}}$: color filter with $K$ pieces, $\alpha$: step size,\\$\epsilon$: parameter bounds}\\

\algorithmicensure{$\boldsymbol{x}'$: adversarial image}
\begin{algorithmic}[1]
\State Initialize $\boldsymbol{\theta}_0\leftarrow \frac{\mathds{1}^K}{K}$, ~$\boldsymbol{x}_0'\leftarrow F_{{\boldsymbol{\theta}_0}}(\boldsymbol{x})$
\For {$t\leftarrow 1$ to $T$}
\State $\boldsymbol{g}\leftarrow \nabla_{\boldsymbol{\theta}}L_{\mathrm{CW}}(\boldsymbol{x}_{t-1}',l)$\Comment{\parbox[t]{.45\linewidth}}{calculate gradients}
\State $\boldsymbol{\theta}_t\leftarrow\boldsymbol{\theta}_{t-1}-\alpha{\frac{\boldsymbol{g}}{{\|\boldsymbol{g}\|}_2}}$\Comment{\parbox[t]{.45\linewidth}}{update parameters}
\State $\boldsymbol{\theta}_{t}\leftarrow \mathrm{clip}(\boldsymbol{\theta}_{t},\frac{1}{K},\frac{\epsilon}{K})$\Comment{\parbox[t]{.45\linewidth}}{clip parameters}
\State $\boldsymbol{x}_{t}'\leftarrow F_{{\boldsymbol{\theta}_{t}}}(\boldsymbol{x})$\Comment{\parbox[t]{.45\linewidth}}{transform the image}
\EndFor
\State \Return $\boldsymbol{x}'\leftarrow\boldsymbol{x}_{t}'$ that is adversarial
\end{algorithmic}
\end{algorithm}

\begin{figure}[!t]
\centering
  \includegraphics[width=\columnwidth]{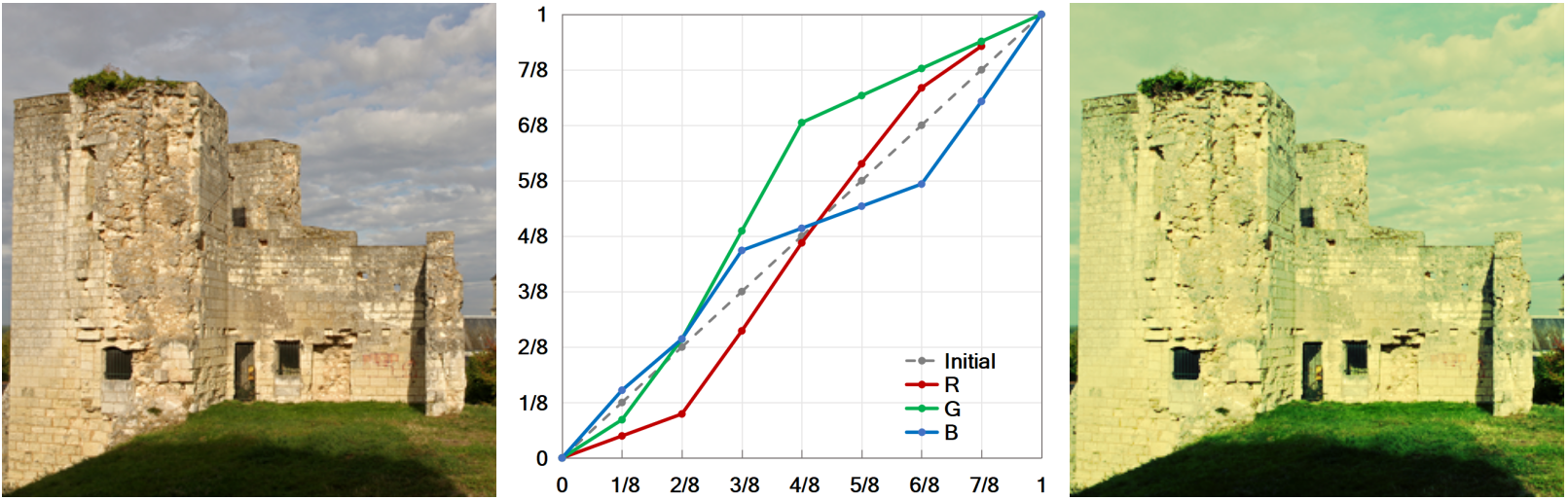}
   \caption{An original image (left) is transformed by using the human-interpretable color curves (middle) in our AdvCF ($K=8$, $\epsilon=4$) to its adversarial version (right).}
\label{fig:curve}
\end{figure}

\subsection{Adversarial Color Filter}
For our new attack, Adversarial Color Filter (AdvCF), we use the above-described color filter~\cite{hu2018exposure} as the basic component.
Specifically, we iteratively optimize the color filter formulated in Eq.~\ref{filter} for achieving non-suspicious adversarial images rather than for the original, image retouching task.
The main novelty of AdvCF over previous color transformation attacks is its explicit color transformation space, which is specified as the parameter space of the color filter with $L_{\infty}$ bounds imposed directly on filter parameters.
AdvCF is updated using PGD, and the parameter bound is set proportional to a specific number of pieces of the color filter.
Note that our parameter bound does not need to be tight since the filter can inherently guarantee color uniformity even when the perturbations are large.
We adopt the well-known C\&W loss (Eq.~\ref{cw2}), and find that the commonly used cross-entropy loss yields similar results.
Note that other attacks, ReColorAdv and DDN, have also mixed C\&W loss with PGD iterative optimization.
The full process of AdvCF is described in Algorithm~\ref{AdvCF}.

Fig.~\ref{fig:curve} gives an example of one successful adversarial image that has been achieved by our AdvCF.
As can be seen, the adversarial image can be obtained by only adjusting the color curves of the original image.
This whole process is directly interpretable to humans because it involves a transformation that is well-known in photography and can be achieved with standard photo editing software.

\subsection{Color-Style-Guided AdvCF}
In order to further boost the image acceptability, we propose to optimize AdvCF towards specific appealing color styles\footnote{We automate the implementation using the GIMP toolkit with the Instagram Effects Plugins provided by \url{https://www.marcocrippa.it/page/gimp_instagram.php}.}.
This new variant can be formulated as:
\begin{equation}
\label{AdvCF_Enh}
\underset{\boldsymbol{\theta}}{\mathrm{min}}~L_{\mathrm{CW}}(F_{\boldsymbol{\theta}}(\boldsymbol{x}),l) +\lambda \cdot{\|F_{\boldsymbol{\theta}}(\boldsymbol{x})-\boldsymbol{x}_{\mathrm{enh}}\|}_2^2,
\end{equation}
where $\boldsymbol{x}_{\mathrm{enh}}$ denotes the target image that has a specific color style.
The optimization still starts from the original image to ensure the whole modification process to be fully modeled by the color curve in our AdvCF.
For the balancing factor $\lambda$, we empirically choose 0.0001 without tuning it via a time-consuming line search.
This is meaningful since we are targeting a holistic color style, compared with targeting minimal pixel perturbations in the C\&W attack.

\begin{figure}[!t]
\begin{center}
  \includegraphics[width=\columnwidth]{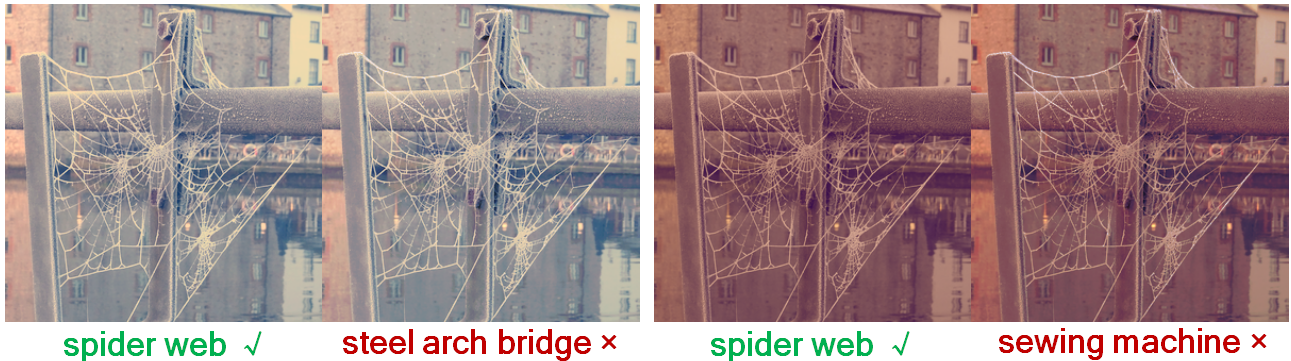}
\end{center}
   \caption{Image examples achieved by popular Instagram filters (left of each pair) and the AdvCF adversarial images guided by the same style (right of each pair).
   Original labels are shown by \textcolor{green}{\cmark} and adversarial labels shown by \textcolor{red}{\xmark}.}
\label{fig:ins}
\end{figure}

Here, specifically, we use popular Instagram filters to specify appealing color styles.
One can also adapt the type of filters to the specific image content by using an automatic filter recommendation technique~\cite{sun2017photo}.
Note that our work explores the adversarial effects and uses Instagram filters only for improving image acceptability, which is different from previous work that uses the effects of Instagram filters to fool the models~\cite{choi2017geo,wu2019recognizing}.
As can be seen from Fig.~\ref{fig:ins} and Table~\ref{tab:ins}, this new variant of AdvCF can achieve appealing color styles while successfully fooling the classifier.
However, solely applying Instagram filters on original images without our AdvCF has little impact on the model accuracy.

\begin{table}[!t]
         \caption{Model accuracy (\%) on filtered images with two popular Instagram styles, Nashville (Insta-N) and Toaster (Insta-T), and on adversarial images created by our AdvCF guided by the same filter styles (AdvCF-N and AdvCF-T).}
\renewcommand{\arraystretch}{1}
      \centering
        \begin{tabular}{c|cc|cc}
\toprule[1pt]
Original&Insta-N&AdvCF-N&Insta-T&AdvCF-T\\
\midrule
95.1&86.2&\textbf{8.7}&87.8&\textbf{8.5}\\

\bottomrule[1pt]
\end{tabular}
\label{tab:ins}
\end{table}

\section{Our New Systematic Robustness Analysis}
\label{sec:analysis}
In this section, we carry out a systematic robustness analysis of image classifiers against adversarial color transformations based on our proposed new attack, AdvCF, from both the attack (Sec.~\ref{sec:ana_attack}) and defense (Sec.~\ref{sec:ana_defense}) perspectives.
This analysis is enabled by the key novelty of AdvCF, i.e., its explicitly specified color filter space.

\begin{figure}[!t]
\centering
  \includegraphics[width=\columnwidth]{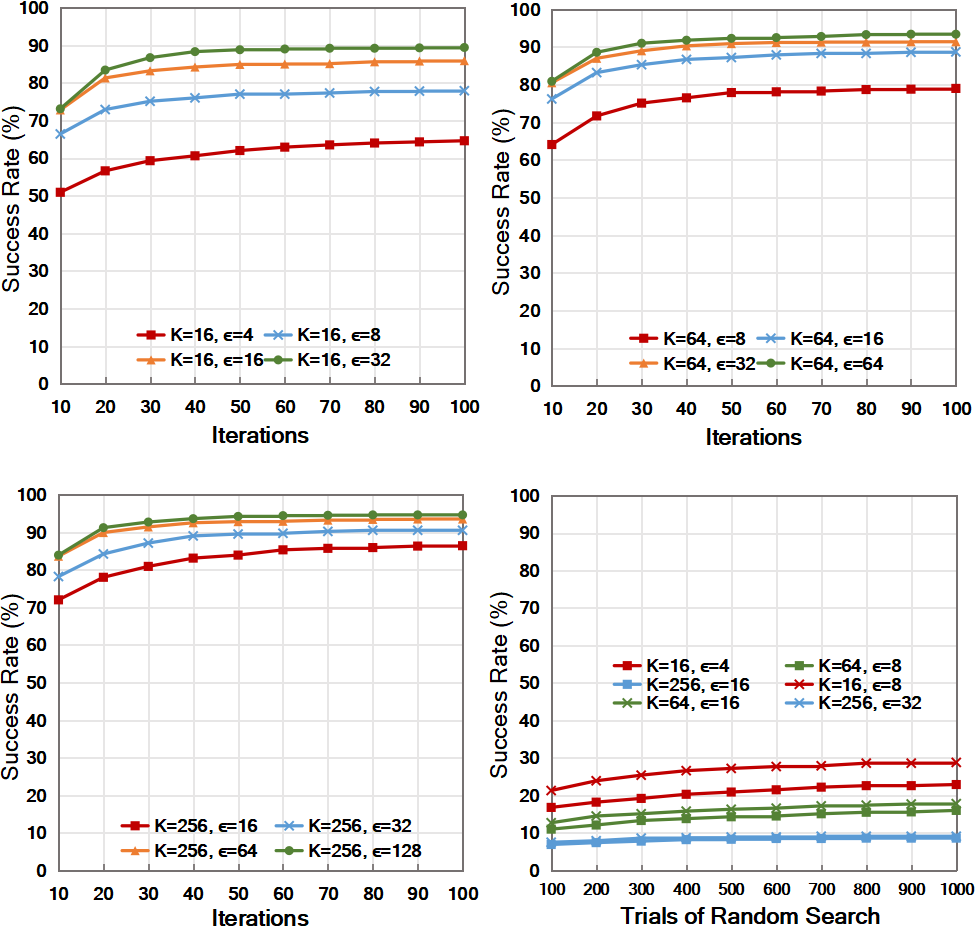}
   \caption{White-box success rates of AdvCF under different parameter bounds $\epsilon$, different number of filter pieces $K$. AdvCF is also compared to a random search-based variant.}
\label{fig:AdvCF_vs_RS}
\end{figure}

\begin{figure}[!t]
\centering
  \includegraphics[width=\columnwidth]{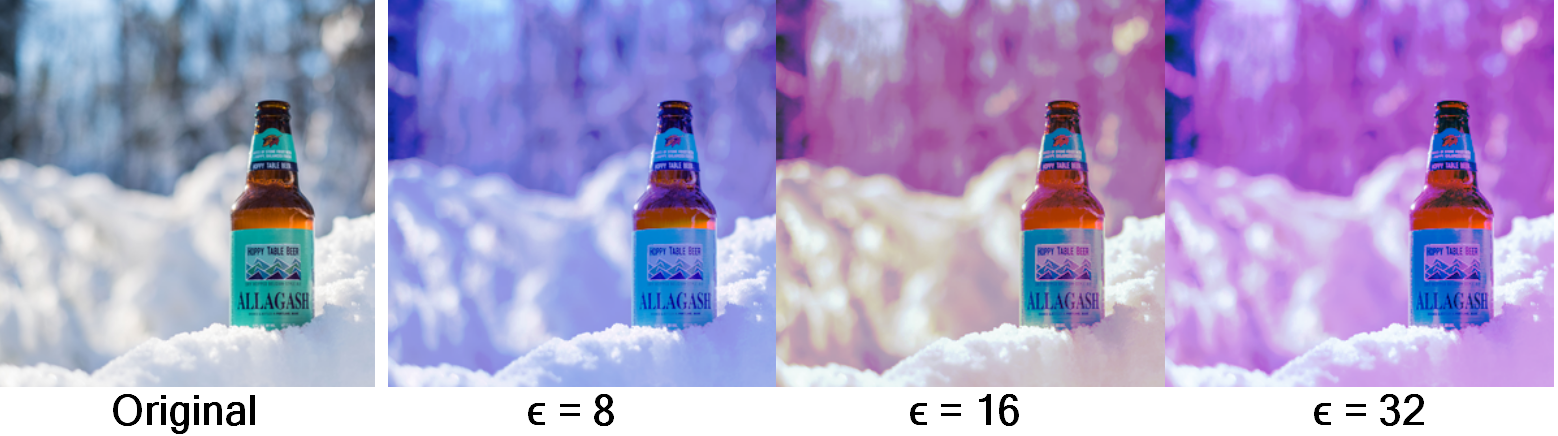}
   \caption{Adversarial images generated by AdvCF with varied parameter bounds, $\epsilon$. $K$ is set as 64.
}
\label{fig:paras_visual}
\end{figure}

\subsection{Analysis from Attack Perspective}
\label{sec:ana_attack}
We conduct our analysis on the ImageNet-Compatible Dataset~\cite{kurakin2018adversarial} and use the Inception-V3 as the white-box model.
This dataset consists of 1000 color images (with the size of 299$\times299$), each of which is associated with one ImageNet class label.
We choose this dataset of natural color images with a relatively large size because we are interested in the visual acceptability of images.

We test AdvCF systematically within its explicit color filter space by varying its parameter bounds $\epsilon$.
As can be seen from Fig.~\ref{fig:AdvCF_vs_RS}, relaxing the parameter bound $\epsilon$ boosts the performance since it allows large color transformations for possible adversarial images.
As can be seen from Fig.~\ref{fig:paras_visual}, the color transformation becomes increasingly stronger as $\epsilon$ is gradually relaxed.
With the same $\epsilon$, a larger number of pieces $K$ leads to better results by allowing more fine-grained color changes but with higher time costs.

We also compare our AdvCF with a random search-based variant, which uniformly samples filter parameters within a range bounded by $\epsilon$.
This variant is conceptually similar to the approaches in ColorFool but rather operated in the filter space instead of the pixel space.
The bottom right sub-figure of Fig.~\ref{fig:AdvCF_vs_RS} shows that even with 10x trials, the adversarial strength of random search is much lower than the original AdvCF under the same $\epsilon$ bound.
Moreover, when applying a higher $K$, the increasing number of parameters makes it even harder to search for possible adversarial images.
In contrast, our AdvCF can be improved by increasing $K$ due to the use of gradient-based optimization.

\subsection{Analysis from Defense Perspective}
\label{sec:ana_defense}
\begin{figure}[!t]
\centering
  \includegraphics[width=\columnwidth]{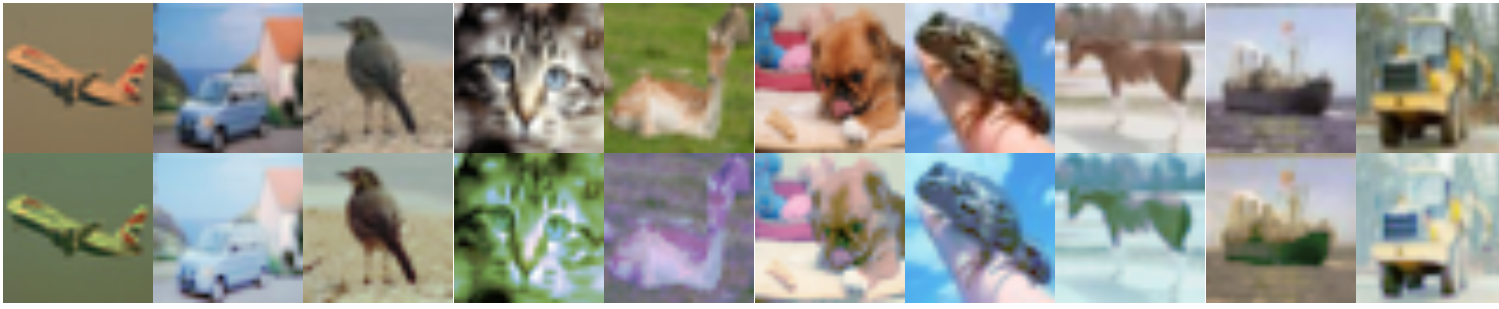}
   \caption{Original images (top) and their adversarial versions (bottom) generated by AdvCF$_{64}^8$ on CIFAR-10.}
\label{fig:exampes_cifar}
\end{figure}

We explore the model robustness against AdvCF when its adversarial images are used for adversarially training the model.
We follow the common practice to use the CIFAR-10 dataset~\cite{krizhevsky2009learning}, since adversarial training on ImageNet remains an open problem~\cite{xie2019feature,xie2019intriguing,shafahi2019adversarial,wong2020fast}. 
The original papers of ColorFool and ReColorAdv also use CIFAR-10 for their adversarial training experiments.
The CIFAR-10 consists of 60000 32$\times$32 color images in 10 object classes.
We use the whole official training split (50000 images) for training, the first 500 of all 10000 testing images for model selection, and the rest 9500 for model evaluation. 
For AdvCF, we use 50 iterations with $\alpha=0.1$, $K=64$, and $\epsilon=8$.
Fig.~\ref{fig:exampes_cifar} visualized some examples of CIFAR-10 adversarial images generated by AdvCF on the original (undefended) model.

For adversarial training, we follow~\cite{madry2017towards} to run 30 epochs and finally select the model that achieves the best standard accuracy on the validation set.
Detailed hyperparameter settings are summarized in Table~IX of Appendix D.
Fig.~\ref{fig:train_curve} (left) shows adversarial training curves for both standard accuracy and robust accuracy.
As can be seen, the adversarially trained model on AdvCF (AT-AdvCF) converges to a high standard accuracy (87.24\%) and a substantial robust accuracy (54.01\%) against the AdvCF attack.
In Fig.~22 of Appendix D, we further visualize the adversarial images generated on AT-AdvCF compared to those on the original model.
As can be seen, AdvCF generally requires stronger transformations for attacking the AT-AdvCF model than attacking the original model.

\begin{figure}[t!]
\centering
  \includegraphics[width=0.47\columnwidth]{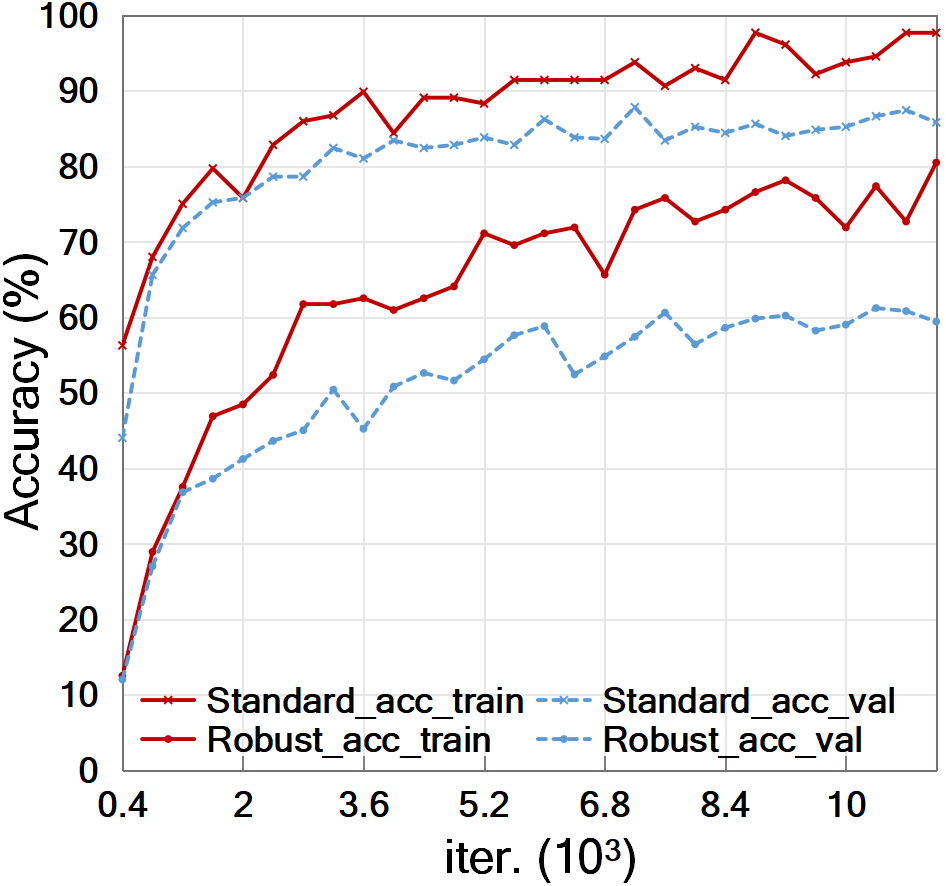}
    \includegraphics[width=0.51\columnwidth,height=0.44\columnwidth]{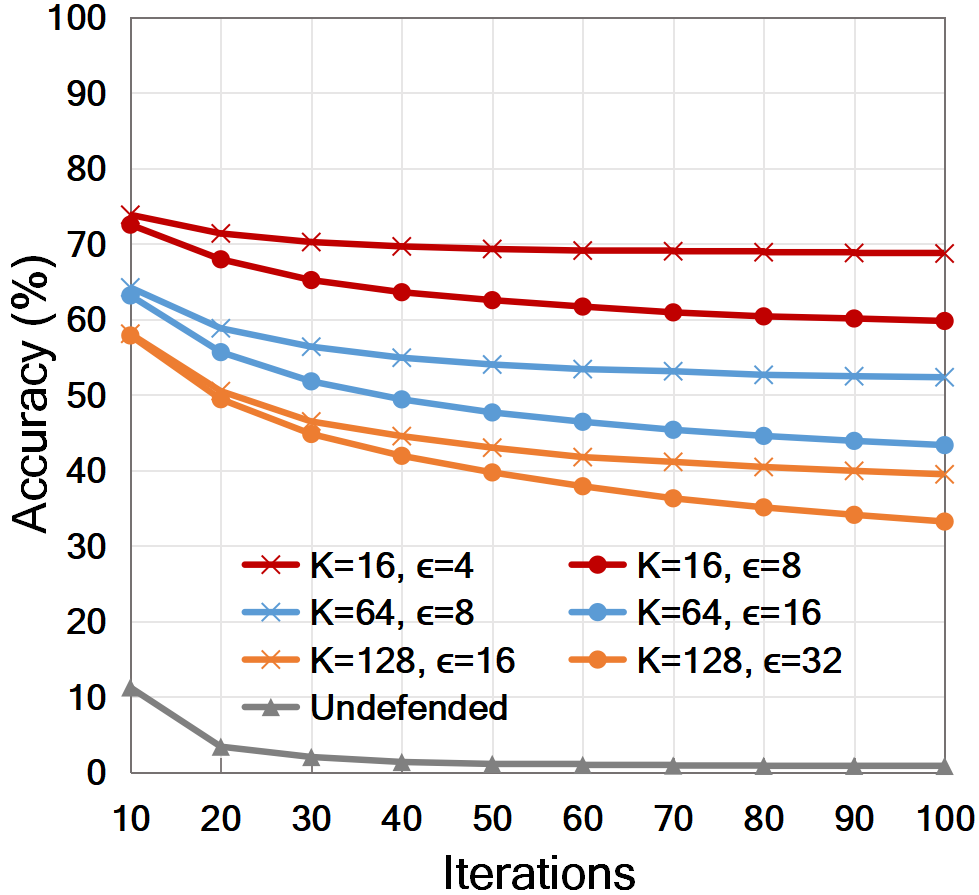}

   \caption{Left: The training curve of our AdvCF adversarial training. Right: Accuracy of the adversarially trained model (on AdvCF$_{64}^8$) tested against AdvCF attacks with varied hyperparameters.}
\label{fig:train_curve}
\end{figure}

We further look at the impact of the parameters of AdvCF on the attack performance against adversarial training.
Fig.~\ref{fig:train_curve} (right) shows the accuracy results when the model is adversarially trained on AdvCF$_{64}^8$ and tested on adversarial images that are generated by AdvCF with varied settings of parameter bounds $\epsilon$ and the number of filter pieces $K$.
As can be seen, with the same $K$, the model is more robust against the adversarial images that have a lower $\epsilon$. 
In addition, under the same $\epsilon$ level, increasing $K$ leads to more fine-grained color transformation, and as a result, it causes a decrease in model robustness.

We also explore the relationship between model robustness against our (non-suspicious, color transformation-based) AdvCF and against conventional indistinguishable adversarial images.
For indistinguishable adversarial images, we follow~\cite{madry2017towards} to use the WideResNet-34-based model and train it on 7-iteration I-FGSM with $\alpha$=2 and $\epsilon=8$ to achieve the $L_{\infty}$-robust model.
As can be seen from Table~\ref{tab:adv_train}, adversarial training substantially improves model robustness in both cases of I-FGSM and AdvCF, but with a decrease in model accuracy on original images. 
We can also observe that the $L_{\infty}$-robustness and our AdvCF-robustness do not cover each other well.
This suggests that achieving comprehensive model robustness should consider different types of adversaries. 

We further evaluate other color transformation attacks against adversarial training.
As can be seen from Table~\ref{tab:adv_train}, in general, AT-AdvCF also improves the model robustness against other two color transformation attacks, ColorFool~\cite{shamsabadi2020colorfool} and ReColorAdv~\cite{Laidlaw2019functional}.
Specifically, AT-AdvCF leads to the highest robustness against ColorFool, which is also based on relatively global color modifications.
In addition, for ReColorAdv, AT-AdvCF also achieves higher robustness than the undefended model.
However, in this case, AT-$L_{\infty}$ leads to the highest robustness.
This observation is consistent with findings in the original ReColorAdv work~\cite{Laidlaw2019functional}, suggesting that its color modifications are indeed more localized than ColorFool and our AdvCF (see our discussion around Fig. 10).
Here we did not consider cAdv because it cannot be used on CIFAR-10 since its deep colorization model only accepts input images with a size of 224×224.
The original work of cAdv also did not consider CIFAR-10.

\begin{table}[!t]
 \caption{Adversarial training (AT) against color transformation attacks vs. I-FGSM.
 AT-$L_{\infty}$: model trained on $L_{\infty}$-bounded I-FGSM, AT-AdvCF: model trained on AdvCF$_{64}^8$.}
\renewcommand{\arraystretch}{1}
\centering
\newcommand{\tabincell}[2]{\begin{tabular}{@{}#1@{}}#2\end{tabular}}
\begin{tabular}{l|ccc}
\toprule[1pt]
Acc (\%)&w/o AT&AT-$L_{\infty}$&AT-AdvCF\\
\midrule
Original&\textbf{95.6}&86.7&87.2\\
I-FGSM&7.8&\textbf{49.6}&5.5\\
\hline
AdvCF&7.1&16.5&\textbf{54.0}\\
ColorFool&7.4&0.7&\textbf{42.4}\\
RecolorAdv&0.2&\textbf{20.1}&8.5\\
\bottomrule[1pt]
\end{tabular}
 \label{tab:adv_train}
\end{table}

\section{Comparing AdvCF with Other Attacks}
\label{sec:compare}
In this section, we compare AdvCF with other representative attacks in terms of both the success rate and visual acceptability of their adversarial images.

Our evaluation is focused on the untargeted attack setting since it is hard for color transformation attacks to achieve targeted misclassification~\cite{hosseini2018semantic,Laidlaw2019functional,shamsabadi2020colorfool}.
In particular, the original studies of RecolorAdv~\cite{Laidlaw2019functional} and Colorfool~\cite{shamsabadi2020colorfool} also only explore untargeted attacks.
Although cAdv~\cite{bhattad2020Unrestricted} tried targeted attacks, the resulting images look unnatural due to the large, localized color perturbations (see examples in their Figure 8).

\subsection{Experimental Settings}
\noindent\textbf{Dataset and models.} We also use the 1000 color images from the ImageNet-Compatible Dataset~\cite{kurakin2018adversarial}.
We consider five diverse classifiers that are pre-trained on ImageNet: Inception-V3~\cite{szegedy2016rethinking}, AlexNet~\cite{krizhevsky2012imagenet}, ResNet50~\cite{he2016deep}, VGG19~\cite{simonyan2014very}, and DenseNet~\cite{huang2017densely}.
If not specifically mentioned, we adopt Inception-V3 as the white-box source model because it is the official model used in the NIPS 2017 Competition.

\noindent\textbf{Baselines.} We evaluate our AdvCF and other color transformation attacks we reviewed in Sec.~\ref{sec:non-sus}: ColorFool~\cite{shamsabadi2020colorfool}, ReColorAdv~\cite{Laidlaw2019functional}, and cAdv~\cite{bhattad2020Unrestricted}.
We also consider two indistinguishable attacks: $L_{\infty}$-bounded I-FGSM~\cite{kurakin2016adversarial} and $L_2$-bounded DDN~\cite{rony2019decoupling}.
DDN is based on the well-known C\&W loss but with a PGD optimization for efficiency.
We use 100 iterations of gradient descent for I-FGSM, DDN, ReColorAdv, cAdv, and our AdvCF, and 500 trials for ColorFool since it does not exploit gradient information.
For our AdvCF, if not specifically mentioned, we set the step size $\alpha$ as 1, the number of pieces $K$ as 64, and the parameter bound $\epsilon$ as 16 for a good trade-off between the attack success and image acceptability.

\noindent\textbf{Control of transferability for a fair comparison.} Our evaluation compares a range of attacks that is more diverse than what is considered in conventional literature.
It is not tractable to control visual acceptability because different attacks formulate their image transformations in different ways.
For this reason, we instead control the attack strength (i.e, transferability) and then compare their performance in other aspects, such as run time, attack robustness, perturbation patterns, and image acceptability (through an extensive user study).
In order to ensure a fair comparison, we control different attacks to achieve similar black-box success rates (i.e, transferability), and the results in Table~\ref{tab:trans} confirm that our control is successful. Especially, measuring black-box success also gives more practical implications~\cite{liu2017delving,sun2022exploring,sun2022query}.

Specifically, we first set the black-box success rates of our AdvCF as the target performance, and then gradually increase the constraint levels of other approaches (i.e., $L_{\infty}$ for I-FGSM, $L_{2}$ for DDN, $\epsilon$ in the CIELUV color space for ReColorAdv, and $k$ for cAdv).
This tuning process finally leads to $L_{\infty}=16$ for I-FGSM, $L_{2}=6$ for DDN, $\epsilon=0.18$ in the CIELUV color space, and $k=2$ for cAdv.
Moreover, in order to make sure ReColorAdv and cAdv can converge to optimal performance within 100 iterations, we choose relatively large learning rates (0.05 for ReColorAdv and 0.001 for cAdv) for their optimization. For other hyperparameters, the same values are adopted as in the corresponding original work of the attacks.

\begin{table}[!t]
\caption{Attack performance of different attacks against Inc3 (white-box) and other image classifiers (black-box) on ImageNet-Compatible Dataset.
Run time per image is also reported. All attacks are tuned to have similar black-box success. For our AdvCF, $K=64$ and $\epsilon=16$ are used.}
\newcommand{\tabincell}[2]{\begin{tabular}{@{}#1@{}}#2\end{tabular}}
\renewcommand{\arraystretch}{1}
\begin{center}
\resizebox{\columnwidth}{!}{
\begin{tabular}{l|c|ccccc}
\toprule[1pt]
\multirow{2}{*}{Attack}&Run&\multicolumn{5}{c}{Success Rate (\%)}\\
&time&Inc3$^\ast$&Alex& R50&V19&D121\\
\midrule
I-FGSM~\cite{kurakin2016adversarial}&10s&100.0&30.1&34.5&36.7&33.2\\
DDN~\cite{rony2019decoupling}&10s&100.0&25.9&34.8&36.4&34.6\\
\midrule
cAdv~\cite{bhattad2020Unrestricted}&15s&93.0&58.6&31.2&30.2&34.2\\
ReColorAdv~\cite{Laidlaw2019functional}&15s&90.0&56.5&36.8&37.0&30.3\\
\midrule
ColorFool~\cite{shamsabadi2020colorfool}&40s&60.2&69.6&41.6&27.2&29.3\\
Our AdvCF&12s&88.7&62.1&35.8&29.7&30.3\\
\bottomrule[1pt]
\end{tabular}
    }
\end{center}
\label{tab:trans}
\end{table}

\begin{table}[!t]
         \caption{Success rates (\%) of AdvCF against different image classifiers on ImageNet-Compatible Dataset. Results are measured with respect to the classifiers in the columns when applying AdvCF to the classifiers in the rows.}
\renewcommand{\arraystretch}{1}
      \centering
        \begin{tabular}{l|ccccc}
\toprule[1pt]
&Alex&R50&V19&D121&Inc3\\
\midrule

Alex&\textbf{99.5}&37.8&26.6&23.7&15.5\\
R50&71.0&\textbf{96.8}&34.7&33.1&17.2\\
V19&67.7&40.6&\textbf{95.7}&31.6&16.3\\
D121&69.4&46.8&34.3&\textbf{94.9}&15.9\\
Inc3&62.1&35.8&29.7&30.3&\textbf{88.7}\\
\bottomrule[1pt]
\end{tabular}
         \label{tab:cross_imagenet}
\end{table}

\begin{figure*}[!t]
\begin{center}
  \includegraphics[width=0.9\textwidth]{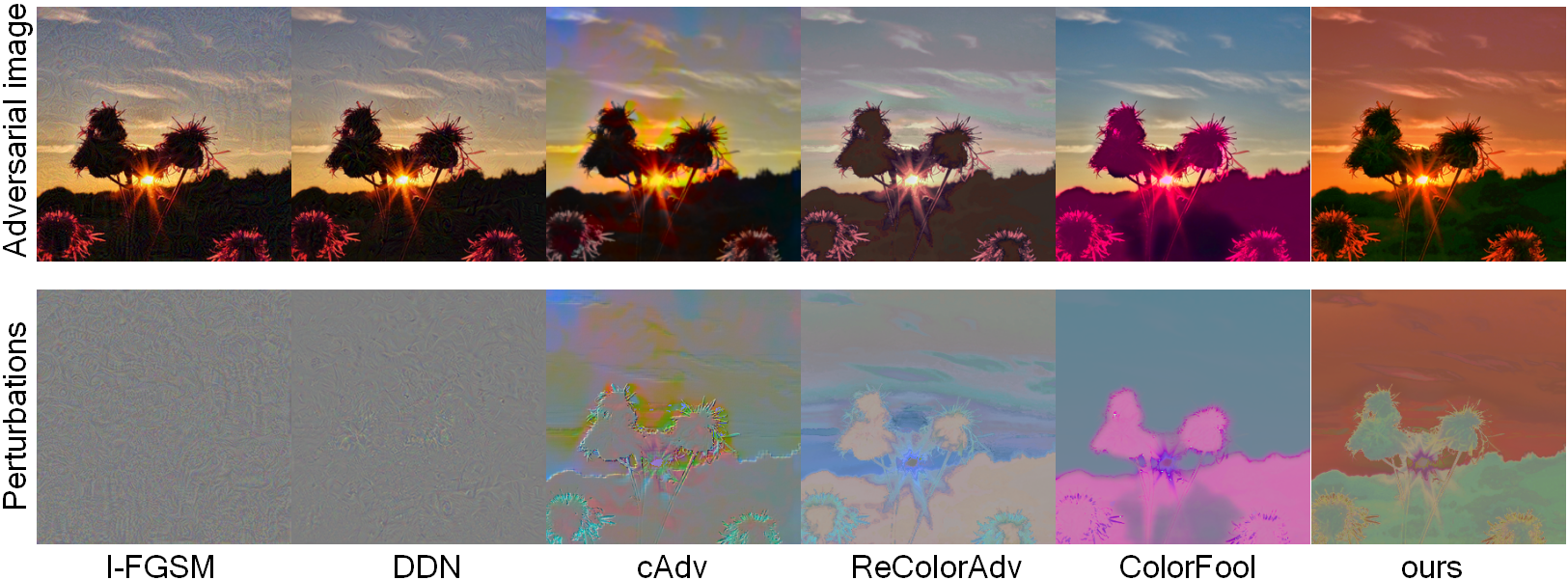}
\end{center}
   \caption{Adversarial images (top) created by different attack approaches have different perturbation patterns (bottom).}
\label{fig:visual}
\end{figure*}

\subsection{Comprehensive Evaluations}

\noindent\textbf{Transferability of AdvCF on other model pairs.} Table~\ref{tab:cross_imagenet} demonstrates the general effectiveness of AdvCF on diverse model architectures.
We find that transferring from a complex architecture (ResNet50, VGG19, DenseNet121, or Inception-V3) to a simple one (AlexNet) is easier to than the other way around.
This finding is consistent with previous work on adversarial transferability~\cite{liu2017delving,shamsabadi2020colorfool,ozbulak2021selection}.
Especially, the Inception-V3 is the most difficult model to attack probably due to its heavily engineered architecture, which contains multiple-size convolution and auxiliary classifiers~\cite{zhao2021success}.

\noindent\textbf{Comparisons on run time.} All the experiments have been run on a single NVIDIA Tesla P100 GPU with 12GB of memory.
As can be seen, our AdvCF is much more efficient than the state-of-the-art human-interpretable approach, ColorFool.
We can also make a trade-off between the success rate and efficiency by adjusting the number of pieces, $K$.
For example, AdvCF with a smaller number of pieces, $K=8$, saves time by 25\% with a decrease of the success rate by 15\%.
AdvCF also supports batch processing for further acceleration.
For example, when being executed in 40 batches of 25 images, the run time is about one second per image.
It is also worth noting that cAdv and ColorFool rely on DNN-based models, which require massive data for pre-training.

\begin{figure}[!t]
\centering
  \includegraphics[width=0.9\columnwidth]{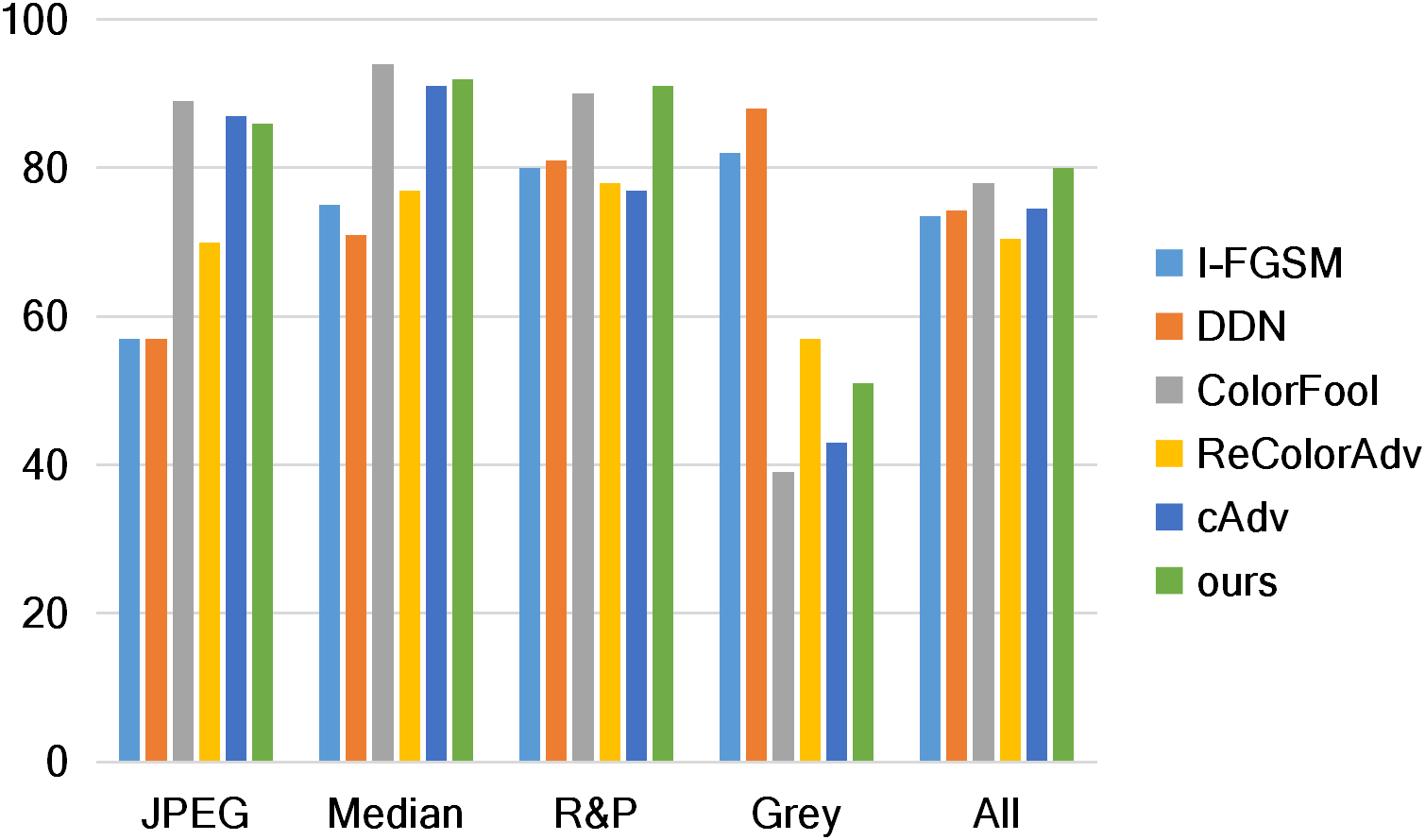}
   \caption{Proportions (\%) of adversarial images that remain successful after input transformations: JPEG compression (Q=30)~\cite{das2018shield}, Median filtering ($3\times3$)~\cite{xu2017feature}, Resizing\&Padding~\cite{xie2017mitigating}, and Gray-scale conversion.
All attacks are with the same parameters as those reported in Table~\ref{tab:trans}. Inc3$\rightarrow$Res50 is shown and other model pairs show similar results (cf. Appendix C).}
\label{fig:input_trans}
\end{figure}

\noindent\textbf{Comparisons on attack robustness.} We compare the robustness of different attacks under defenses that are based on common input transformations.
Existing work has applied common input transformations as defenses against adversarial attacks with little impact on the model performance on original images.
We evaluate different attacks against a wide range of commonly used defensive image transformations:
JPEG compression~\cite{das2018shield,guo2017countering}, median filtering~\cite{xu2017feature}, random resizing\&padding~\cite{xie2017mitigating}.
We also consider gray-scale conversion, which is expected to eliminate the color information used by color transformation.
Specifically, the adversarial images were pre-processed by input transformations before fed into the target model.
For each target model, we calculate the proportion of adversarial images that remain successful after the input transformations.

As can be seen from Fig.~\ref{fig:input_trans}, our AdvCF achieves the best overall performance.
The two indistinguishable attacks (I-FGSM and DDN) are more vulnerable to the three commonly used input transformations, especially JPEG compression and median filtering.
In contrast, the four color transformation attacks (cAdv, ReColorAdv, ColorFool, and our AdvCF) are more robust to these transformations but more vulnerable to gray-scale conversion.
Among these color transformation attacks, ReColorAdv performs best against the gray-scale conversion, probably because it relies on modifications of brightness more than colors during optimizing the adversarial images.

\noindent\textbf{Comparisons on perturbation patterns.} Fig.~\ref{fig:visual} visualizes perturbation patterns of the adversarial images achieved by different attacks.
As can be seen, I-FGSM and DDN generate high-frequency perturbations.
cAdv and ReColorAdv yield more structured perturbations with local colorization artifacts.
Differently, ColorFool and AdvCF images have more homogenous colors.

\begin{table}[!t]
         \caption{Results of the user study on visual acceptability of adversarial images generated by four color transformation attacks.
         The table contains one row per attack and shows the percentage (\%) of times this attack was chosen over the attacks in the columns.}
\renewcommand{\arraystretch}{1}
      \centering
        \begin{tabular}{l|cccc}
\toprule[1pt]
&ReColorAdv&cAdv&ColorFool&Ours\\
\midrule
ReColorAdv&-&44.3&69.7&67.1\\
cAdv&55.7&-&74.0&68.0\\
\midrule
ColorFool&30.3&26.0&-&39.9\\
Ours&32.9&32.0&60.1&-\\
\bottomrule[1pt]
\end{tabular}
\label{tab:user}
\end{table}

\noindent\textbf{Comparisons on image acceptability.} We conduct a user study on Amazon Mechanical Turk, in order to compare the visual acceptability of adversarial images that are created by the four color transformation attacks.
Some previous work~\cite{xiao2018spatially,bhattad2020Unrestricted} has also conducted user studies on adversarial images but is limited to distinguishability, i.e., whether or not an adversarial image can be distinguished from its original version.
Differently, we move beyond distinguishability and assess the relative level of image acceptability with respect to common use scenarios.
We adopt a set of 222 image examples comprised of all the images for which the four attacks can succeed on any of the four black-box models.
All images are judged at the size of 224X224.
Since it is hard to directly decide whether an image is visually acceptable or not, we ask the workers to judge two attacks in pairs with a side-by-side comparison.
We present each image example in a pyramid of three images, with the original image at the top, and two adversarial images generated by two different attacks on the left and right at the bottom (cf. Fig.~18 in Appendix B).
We show 10 different image pyramids and pay 0.4 dollars per task.

\begin{figure}[!t]
\begin{center}
  \includegraphics[width=1\columnwidth]{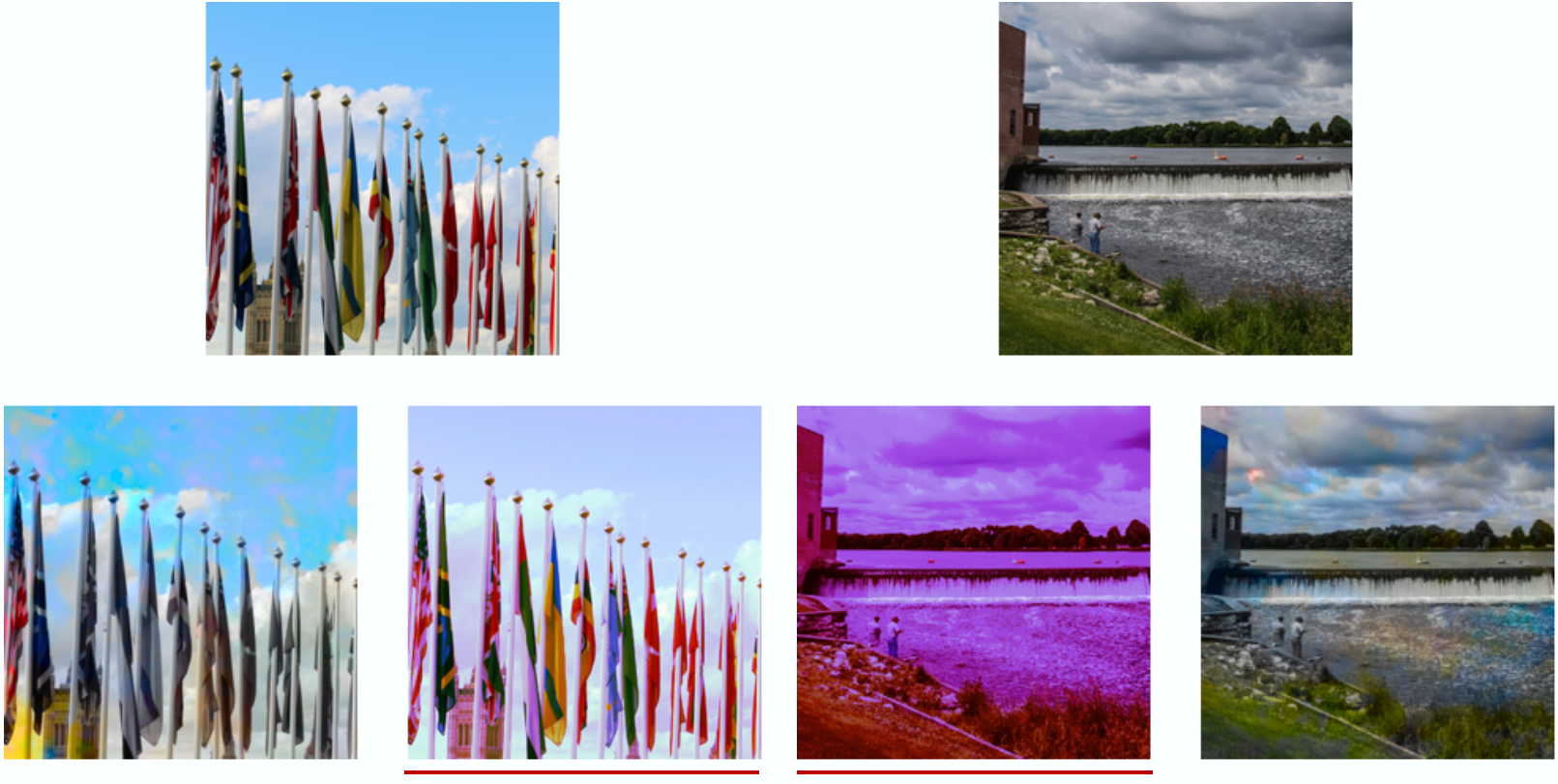}
\end{center}
   \caption{Examples of cases in which our AdvCF was consistently (with at least 8 of 10 workers agreed) chosen to be more (left) or less (right) visually acceptable. The images with red underlines are by AdvCF and the others are by cAdv. Additional examples can be found in Appendix B.
   }
\label{fig:fail}
\end{figure}

We ask the workers to imagine a common use scenario of sharing an image on some social media platform, such as Facebook.
In this scenario, they need to choose which of the two adversarial images they find more visually acceptable for sharing.
We inject random quality control cases, consisting of an original image and two identical adversarial images, to make sure the results are reliable.
For each of the 222 image examples, we only select the judgments from the 10 most reliable workers who have the highest F-score (beta=0.5) on these quality control examples.
This leads to 262 workers selected from all the 346 participating workers. 
On average these workers achieve an F-score of 0.97 (precision as 0.98 and recall as 0.93) on the quality control cases.

The user study results are reported in Table~\ref{tab:user}.
As can be seen, our AdvCF was chosen 60.1\% of the time when compared to the other human-interpretable approach, ColorFool.
In other cases involving the two non-interpretable approaches, AdvCF yields better performance than ColorFool.
Our further manual inspections show that ColorFool usually results in unrealistic colors on the visual concepts that are not specifically constrained.
In addition, segmentation artifacts are also noticed since the semantic segmentation model can not perfectly segment different concepts.

We can also observe that the users have thought the images from ReColorAdv and cAdv to be more visually acceptable than our AdvCF.
The superior performance of ReColorAdv is easily understood because its optimization has applied stronger constraints to the color channels than to the luminance channels based on the fact that human vision is more sensitive to changes in colors than in luminance~\cite{Croce_2019_ICCV,Laidlaw2019functional}.
This results in images like faded pictures (cf. Fig.~\ref{fig:visual} as well as Fig.~19 and Fig.~20 in Appendix B).
For cAdv vs. AdvCF, we visualize some image examples from our user study in Fig.~\ref{fig:fail}.
As can be seen, cAdv results in more localized color modifications due to the use of sparse color hints. 
In this case, AdvCF is more acceptable when the colors in AdvCF images look natural with respect to the image semantics.
In contrast, when the AdvCF images contain semantics with unnatural colors, cAdv is more acceptable.

\section{Experiments in Other Visual Tasks}
\label{sec:other}
In this section, we conduct additional experiments to gain insight into understanding the model robustness against AdvCF in other tasks, including scene recognition, semantic segmentation, and another previously unexplored but relevant task, aesthetic quality assessment.
It is important to consider larger-scale datasets or more diverse tasks for evaluating adversarial attacks because their performance may be biased to the specifically selected images~\cite{ozbulak2021selection}.

\begin{table}[!t]
         \caption{Success rates (\%) of AdvCF in scene recognition.}
\renewcommand{\arraystretch}{1}
      \centering
        \begin{tabular}{l|cccc}
\toprule[1pt]
&Alex&R18&R50&D161\\
\midrule
Alex&\textbf{99.8}&49.0&41.5&38.3\\
R18&79.0&\textbf{98.8}&54.0&47.8\\
R50&76.5&63.6&\textbf{98.7}&51.2\\
D161&79.1&59.0&59.6&\textbf{98.0}\\
\bottomrule[1pt]
\end{tabular}
         \label{tab:cross_scene}
\end{table}

\subsection{Scene Recognition}
Scene recognition is challenging since it involves understanding the characteristic concepts in different scene categories.
We consider a specific task introduced by the Pixel Privacy Challenge~\cite{ppoverview2018} at the MediaEval Multimedia Benchmark.
This task is aimed to protect users' privacy reflected in their scene images shared online and has been explored in previous work on adversarial images~\cite{li2019scene,sanchez2020exploiting,shamsabadi2020colorfool}.
The color transformation achieved by our AdvCF is particularly suitable for this task since they are generated in a way that is aligned with common practices for photo enhancement.
We use the official development set containing 600 images from 60 privacy-sensitive scene categories extracted from the large-scale Places dataset~\cite{zhou2017places}.
All the four classifiers (AlexNet, ResNet18, ResNet50, and DenseNet161) provided by~\cite{zhou2017places} were tested.

As can be seen from Table~\ref{tab:cross_scene}, our AdvCF effectively fools the scene recognition models in both the white-box and black-box scenarios.
When comparing these results with those in Table~\ref{tab:cross_imagenet}, we find that scene recognition is generally easier to attack than ImageNet classification.
This might be due to the fact that the top predictions of scenes generally have lower confidence than those of objects because scenes tend to share more concepts with each other~\cite{cheng2018scene,zhao2018volcano}.

\subsection{Semantic Segmentation} 
Semantic segmentation can be regarded as a pixel-level classification process, i.e., linking each pixel in an image to a class label.
Existing work on attacking semantic segmentation algorithms has been focused on imperceptible perturbations~\cite{arnab2018robustness,xie2017adversarial}.
Here we study the vulnerability of semantic segmentation algorithms to color transformation for the first time.
To this end, the loss function of the iterative optimization of AdvCF can be modified to:
\begin{equation}
L= J(\mathcal{S}(F_{\boldsymbol{\theta}}(\boldsymbol{x})),l^{M\times N}),
\end{equation}
where $\mathcal{S}$ is the segmentation model, $J(\cdot,\cdot)$ is the cross-entropy loss, and the $l^{M\times N}$ is the 2-dimensional ground-truth label map with the same size as the input image.

\begin{table}[!t]
 \caption{Attack performance of AdvCF in semantic segmentation. The mean IoU Ratio (mIoU-R) is reported (lower means stronger attacks). We consider Deeplab-v3~\cite{chen2017rethinking} with ResNet50 (DL-50) and ResNet101 (DL-101), PSPNet~\cite{zhao2017pyramid} with ResNet50 (PSP-50) and ResNet101 (PSP-101).}
\renewcommand{\arraystretch}{1}
\centering
\newcommand{\tabincell}[2]{\begin{tabular}{@{}#1@{}}#2\end{tabular}}
\begin{tabular}{l|cccc}
\toprule[1pt]
mIoU-R (\%)&DL-50&DL-101&PSP-50&PSP-101\\
\midrule[1pt]
DL-50&\textbf{44.8}&73.8&60.9&62.4\\
DL-101&76.6&\textbf{44.6}&64.4&65.3\\
PSP-50&74.9&74.8&\textbf{27.6}&47.8\\
PSP-101&73.1&70.9&44.2&\textbf{31.0}\\
\bottomrule[1pt]
\end{tabular}
 \label{tab:seg}
\end{table}

We use the official validation set (1449 images) of the Pascal VOC dataset~\cite{everingham2015pascal}, which consists of internet-images labeled with 21 different object classes.
We consider two well-known state-of-the-art semantic segmentation algorithms: Deeplab-v3~\cite{chen2017rethinking} and PSPNet~\cite{zhao2017pyramid}, and chose two different model architectures (ResNet-50 and ResNet-101) for each algorithm.
Following~\cite{arnab2018robustness}, we report the mean Intersection over Union (mIoU) Ratio – the ratio of
the mIoU on adversarial images to that on original images.

As can be seen from Table~\ref{tab:seg}, our AdvCF effectively fools the models in the white-box scenario, and such fooling effects can transfer across different architectures and algorithms.
Specifically, the PSPNet is generally less robust than Deeplab-v3 to our AdvCF, despite its higher mIoU on original images (73.94\% vs. 68.64\% with ResNet-50).
This difference may be because the dilated convolution in Deeplab-v3 can learn features at multiple scales, making it harder for the attacks to find a global solution.

\begin{figure}[!t]
\centering
  \includegraphics[width=0.7\columnwidth]{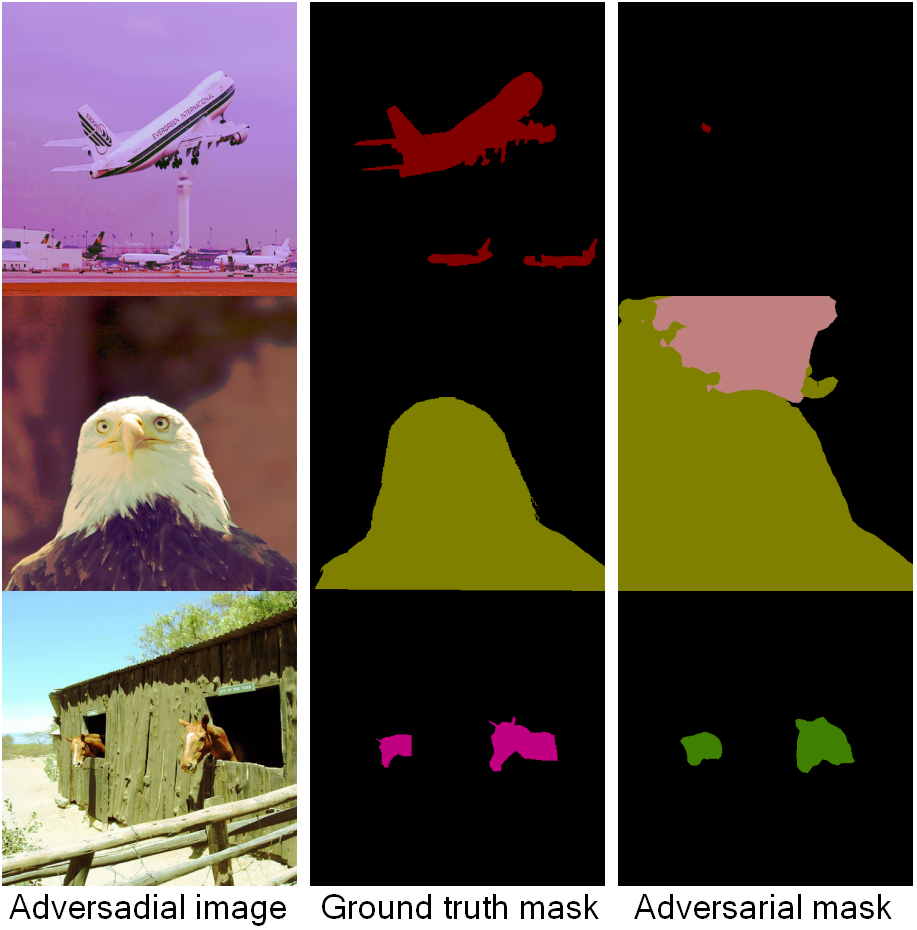}
   \caption{Adversarial images generated by AdvCF in semantic segmentation. PSPNet-ResNet50 is used as the target model. Different colors denote different object classes.}
\label{fig:seg}
\end{figure}

As shown in Fig.~\ref{fig:seg}, the predicted segmentation maps corrupt along human-interpretable dimensions and can be categorized into, from top to bottom, hiding the foreground objects (false negative), disturbing the background (false positive), and misclassifying the objects.
This is essentially different from the irregular corruptions caused by conventional imperceptible perturbations as studied in previous work~\cite{arnab2018robustness,xie2017adversarial}.

\subsection{Aesthetic Quality Assessment}
Aesthetic Quality Assessment (AQA) has become increasingly useful with the prevalence of social multimedia.
Nonetheless, to the best of our knowledge, until now, there is no research on understanding the vulnerability of AQA against adversarial image examples.
We fill this gap by considering an aesthetic quality binary classification task, which has been commonly explored in the literature~\cite{lu2015deep,mai2016composition,talebi2018nima}.
In this task, images with mean ratings smaller or equal to 5 are labeled as low quality and those with mean ratings larger than 5 are labeled as high quality.
We randomly select a subset of 1000 high-quality images from the AVA dataset~\cite{murray2012ava} for our experiments and measure the success rate of the attacks on it.
The well-known NIMA (Neural IMage Assessment)~\cite{talebi2018nima} was adopted as the target AQA model, which can output a predicted quality score for any given image.

The goal of the adversarial images in this task is to cause an AQA model to misclassify high-quality images as low-quality images.
The motivating use scenario is that high-quality images shared online can be misappropriated and misused, e.g., for commercial advertising, in violation of intellectual property rights. 
We can prevent high-quality images from being promoted by fooling the AQA model to misclassify them as low-quality images.

Concretely, for generating adversarial images in this task, the loss function of AdvCF is modified to:
\begin{equation}
L=\mathrm{max}(\textrm{NIMA}(F_{\boldsymbol{\theta}}(\boldsymbol{x})),T),
\end{equation}
where the optimization will be stopped once the output score becomes lower than the pre-defined threshold $T$ (here, 5).
In order to ensure that the actual aesthetics of adversarial images are not decreased compared to the original images, here we adopt our color-style-guided variants of AdvCF, AdvCF-N and AdvCF-T (Eq.~\ref{AdvCF_Enh}).
The very low success rates for Insta-T and Insta-N confirm that the actual aesthetics of images are not decreased when only the appealing Instagram styles are applied without adversarial effects.

Table~\ref{tab:NIMA} shows the attack results. As can be seen, NIMA model is much more robust against AdvCF-N and AdvCF-T than against I-FGSM and DDN.
We also notice that even under $L_{\infty}=1$, I-FGSM and DDN still achieve the same perfect performance.
This contrast suggests that the NIMA model has learned robust features corresponding to relatively uniform, color transformations~\cite{talebi2018nima}, and when the color transformation attacks aim to modify such features, it is challenging for them to succeed.
However, I-FGSM and DDN, which are based on additive pixel perturbations, are still able to succeed by modifying the non-robust features~\cite{ilyas2019adversarial}.

\begin{table}[!t]
     \caption{Attack success rates (\%) in binary aesthetic quality classification. For comparison, the results of Instagram-filtered images (Insta-N and Insta-T) are also reported.}
\renewcommand{\arraystretch}{1}
\centering
\resizebox{\columnwidth}{!}{
\begin{tabular}{cccccc}
\toprule[1pt]
I-FGSM&DDN&Insta-N&Insta-T&AdvCF-N&AdvCF-T\\
\midrule
100.0&100.0&1.7&1.6&25.6&39.3\\
\bottomrule[1pt]
\end{tabular}
}
\label{tab:NIMA}
\end{table}

\section{Conclusion}
In this paper, we have proposed Adversarial Color Filter (AdvCF), a novel color transformation attack that makes it possible to achieve non-suspicious adversarial images by optimizing a simple color filter.
The main novelty of AdvCF is its explicitly specified color filter space, which enables our systematic robustness analysis of adversarial color transformations from both the attack and defense perspectives.
Moreover, AdvCF is human-interpretable since it mimics the color curve adjustment widely applied in the photo retouching process.

We conduct comprehensive comparisons between our AdvCF and another three recent color transformation attacks, and in particular, we provide a user study of the visual acceptability of their created adversarial images.
We also provide interesting insights into understanding the model robustness against AdvCF in other visual tasks, such as scene recognition, semantic segmentation, and aesthetic quality assessment.

We expect AdvCF to be further improved by applying color-adaptive optimization, and concept-adaptive constraints as done in~\cite{shamsabadi2020colorfool}. 
We also expect the robustness against adversarial color transformations, including our AdvCF, will be further improved based on the analysis and findings presented here.
Another interesting direction is to explore the possibility of using AdvCF-based adversarial training for improving robustness against color variations that naturally occur in the real world.

\ifCLASSOPTIONcompsoc
  \section*{Acknowledgments}
  
\else
  \section*{Acknowledgment}
\fi

This work was partially carried out on the Dutch national e-infrastructure with the support of SURF Cooperative.
We would also like to thank the workers participating in our user study on Amazon Mechanical Turk.
\ifCLASSOPTIONcaptionsoff
  \newpage
\fi

\bibliographystyle{IEEEtran}
\bibliography{references}

 \clearpage
\newpage
\appendices

\section{Additional Adversarial Images by AdvCF}
\label{app:img}



\noindent\begin{minipage}{\textwidth}
\centering
\includegraphics[width=0.9\textwidth]{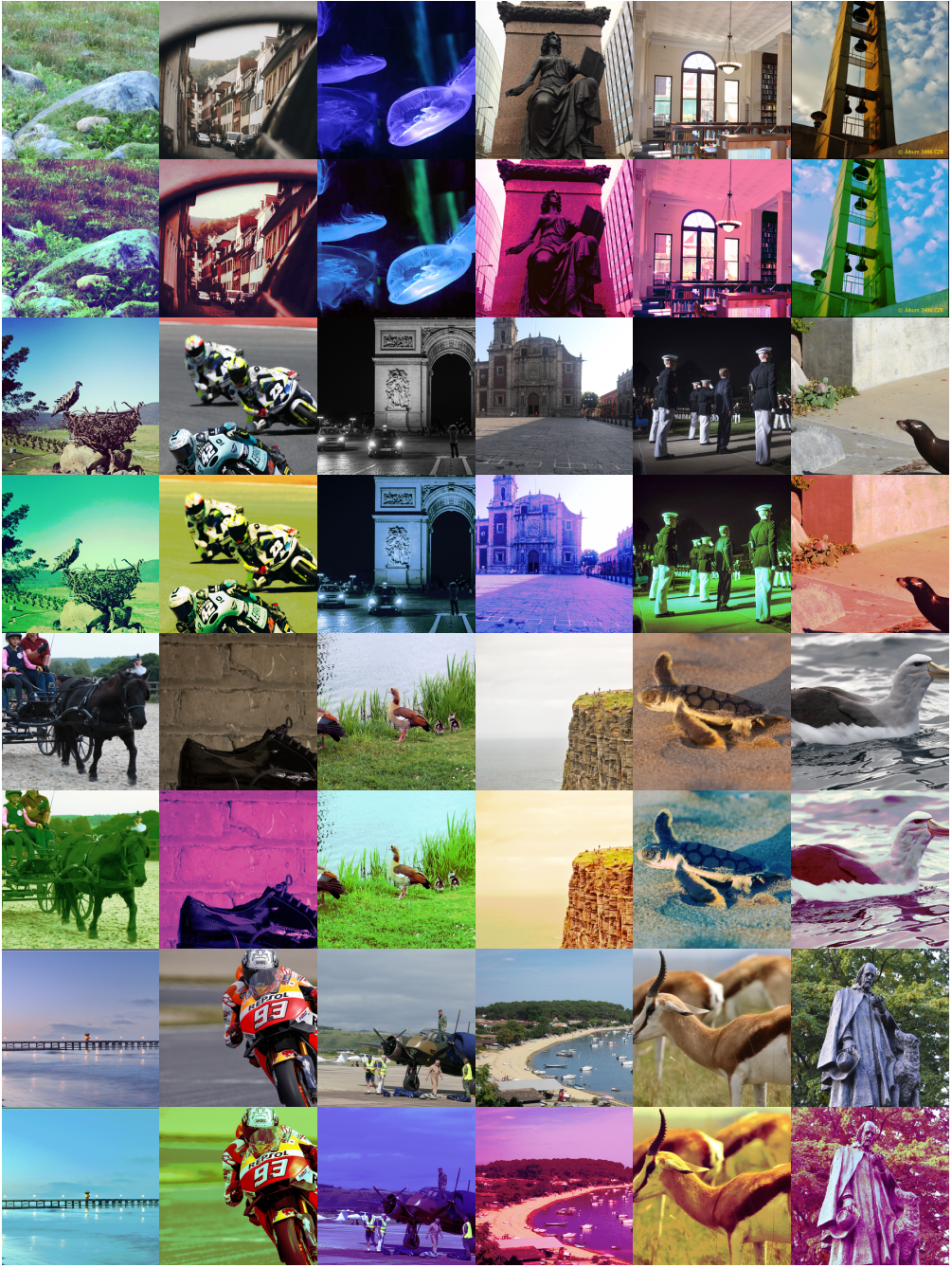}
\captionof{figure}{Successful adversarial images on ImageNet classification with their original versions.}

\label{fig:imagenet_1}
\end{minipage}

 \clearpage
\newpage

\begin{figure*}[!b]
\centering
  \includegraphics[width=\textwidth]{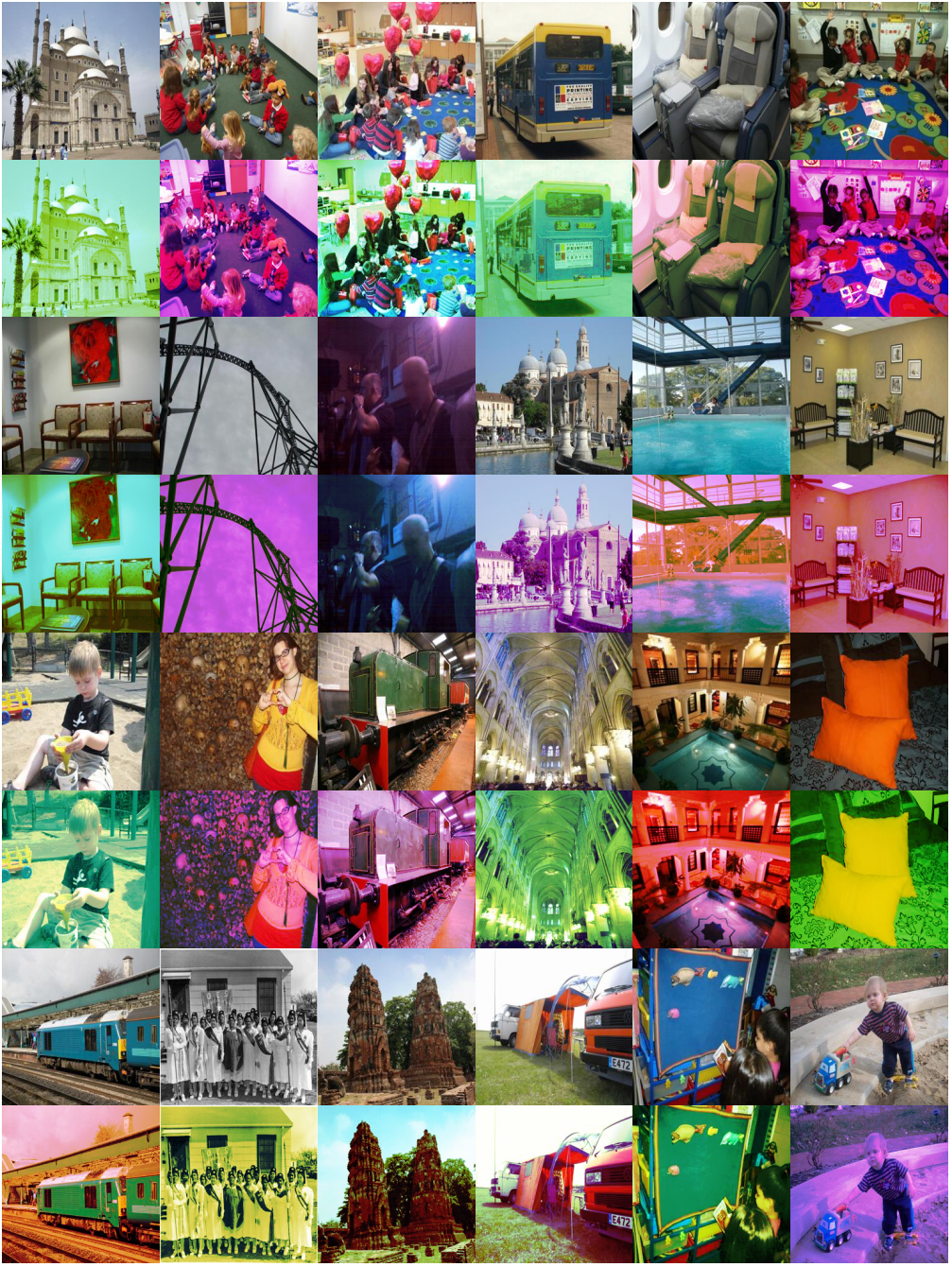}
  \caption{Successful adversarial images on Places (scene recognition) with their original versions.}
\label{fig:scene_1}
\vspace{-0.2cm}
\end{figure*}

\begin{figure*}[!b]
\centering
  \includegraphics[width=\textwidth]{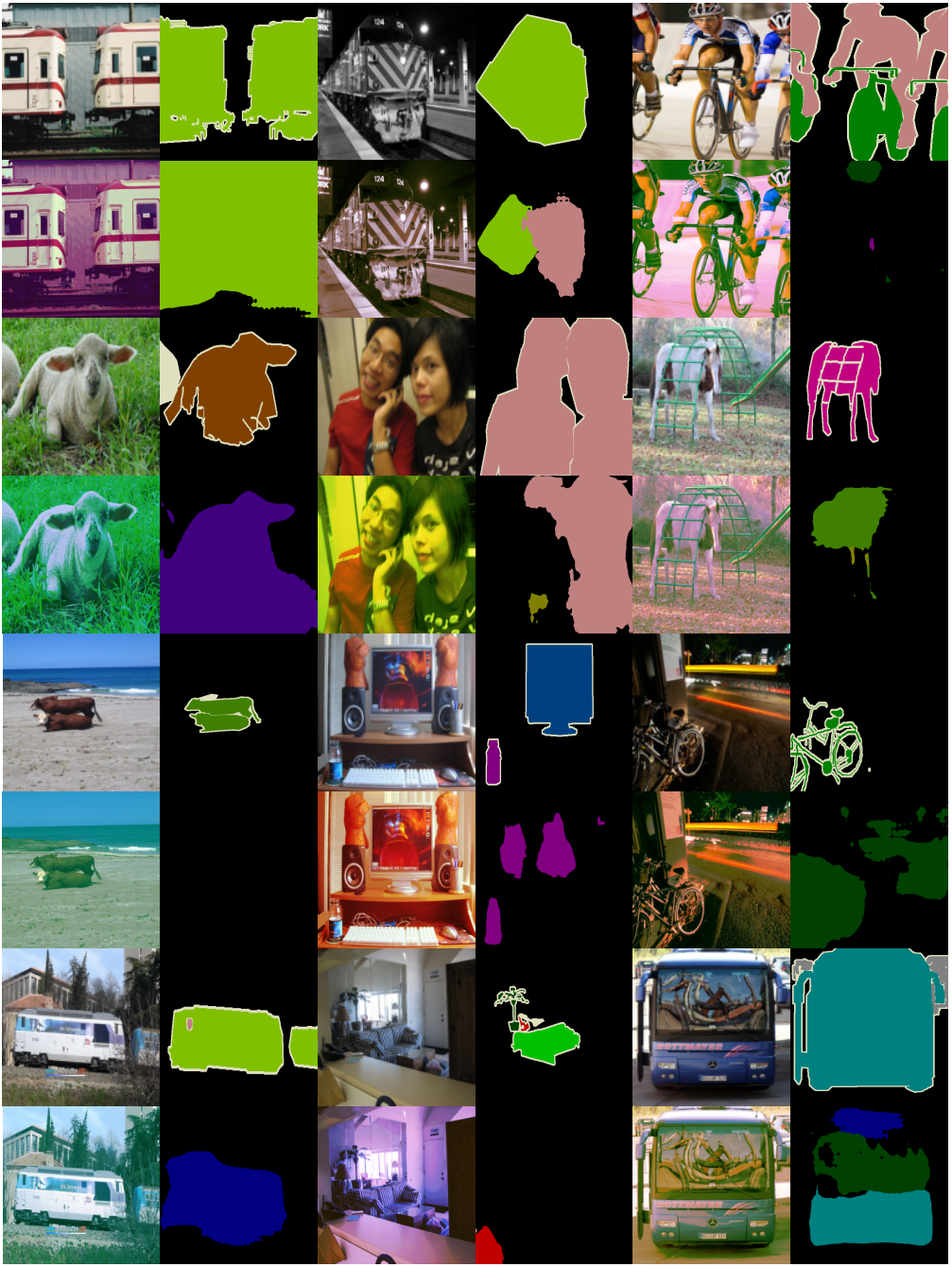}
  \caption{Successful adversarial images on Pascal VOC (semantic segmentation) with their original versions.}
\label{fig:seg_1}
\vspace{-0.2cm}
\end{figure*}

\begin{figure*}[!b]
\centering
  \includegraphics[width=\textwidth]{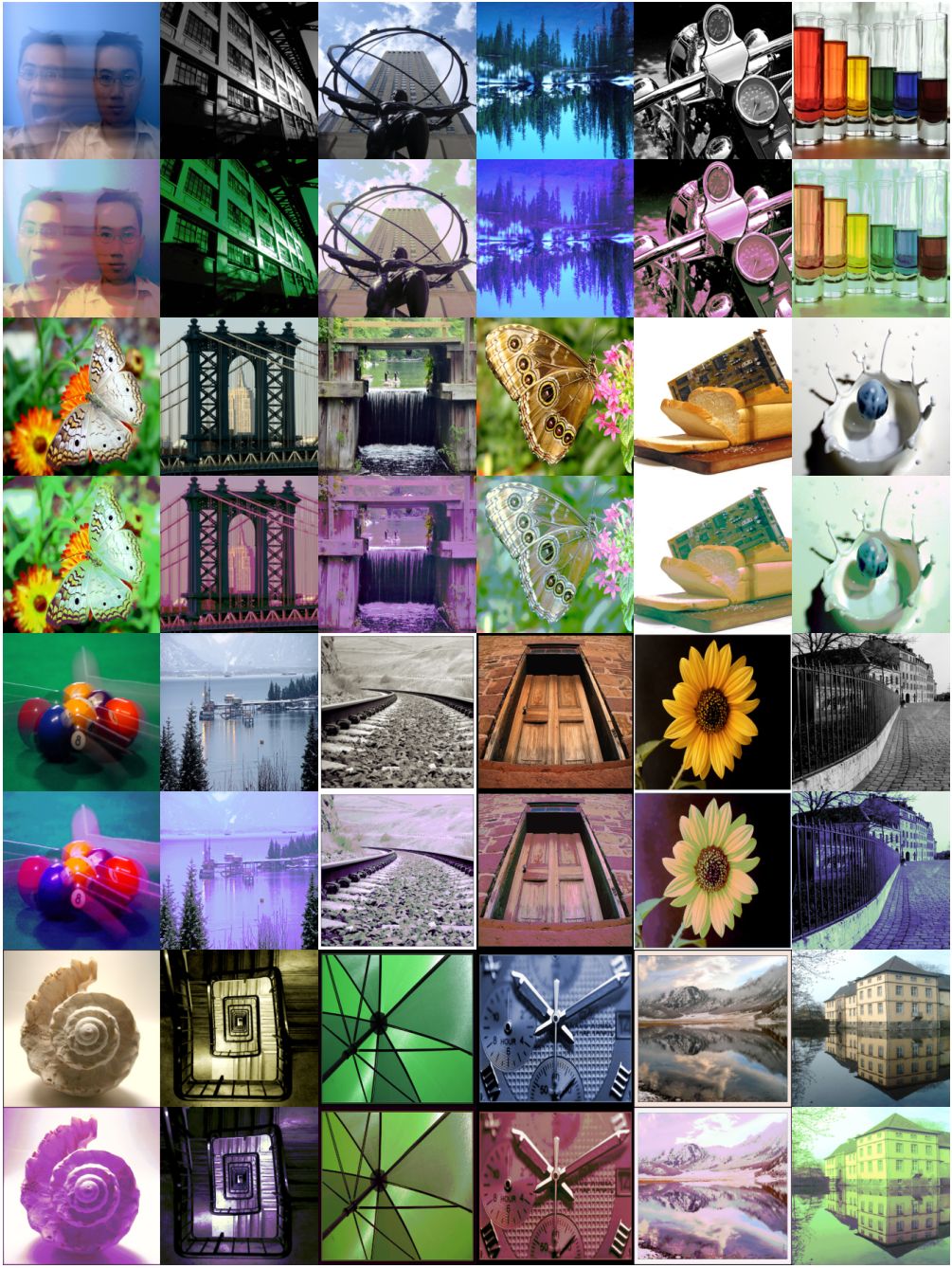}
  \caption{Successful adversarial images on AVA (aesthetics prediction) with their original versions.}
\label{fig:aesthetics_1}
\vspace{-0.2cm}
\end{figure*}

\begin{figure*}[!b]
\centering
  \includegraphics[width=0.95\textwidth]{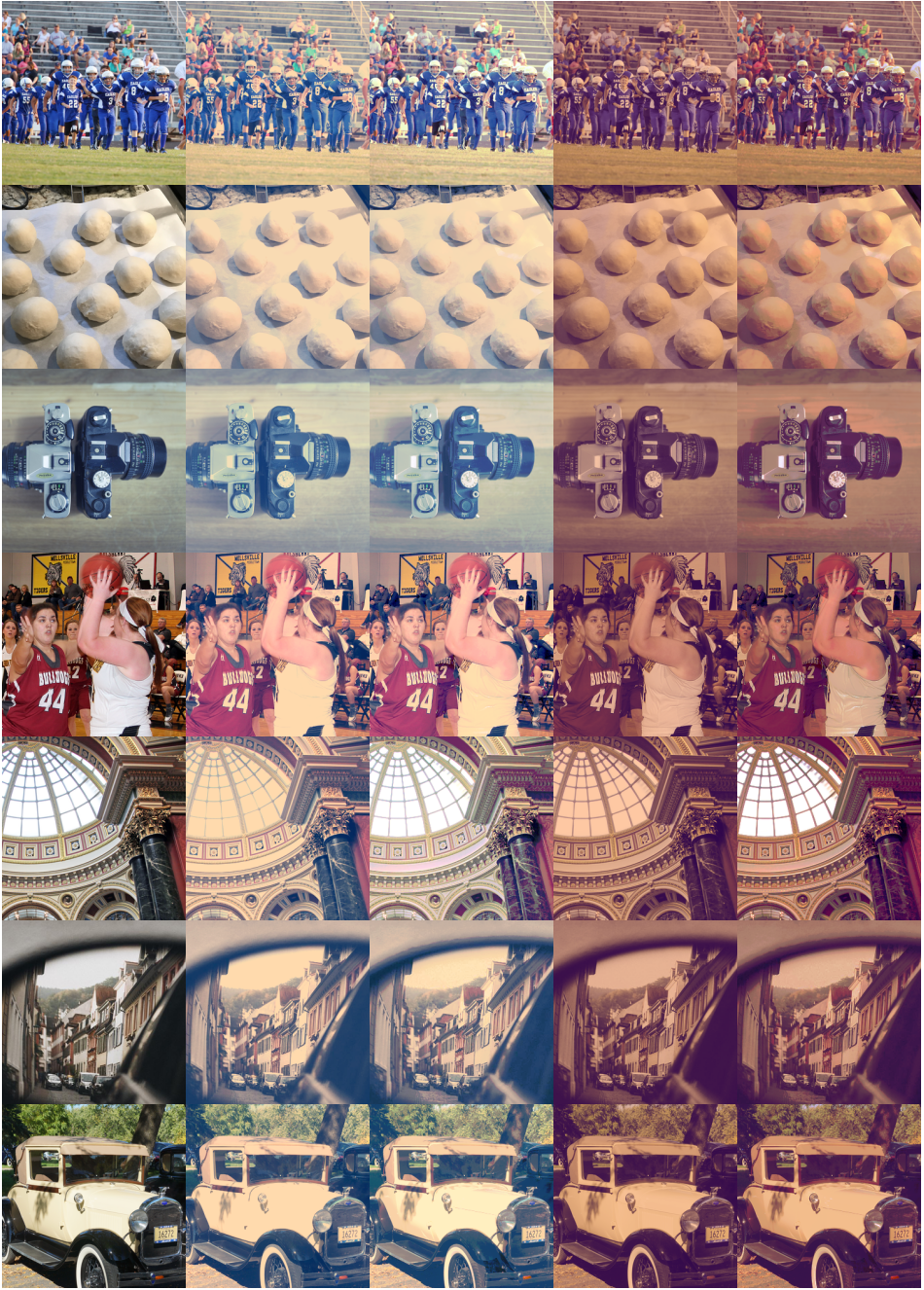}
  \caption{Successful adversarial images achieved by AdvCF targeting particular appealing color styles from Instagram filters. The left column shows the original images. For each pair of filtered images, the left shows the one achieved by one Instagram filter, and the right shows the successful adversarial images created by AdvCF targeting the same style.}
\label{fig:Ins_1}
\vspace{-0.2cm}
\end{figure*}

 \clearpage
\newpage

\section{User Study}
\label{app:user}

\noindent\begin{minipage}{\textwidth}
\centering
\includegraphics[width=0.9\textwidth]{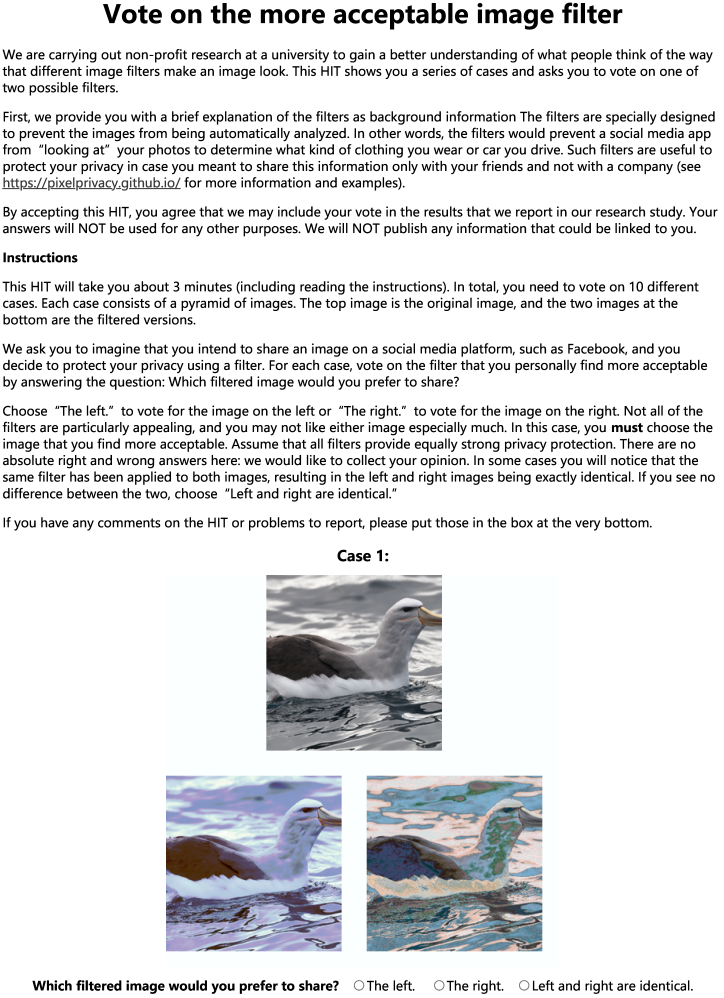}
\captionof{figure}{A screen shot with the exactly the same content used for our user study.}
\label{fig:app_user}
\end{minipage}

\clearpage
\newpage

\noindent\begin{minipage}{\textwidth}
\centering
\includegraphics[width=0.95\textwidth]{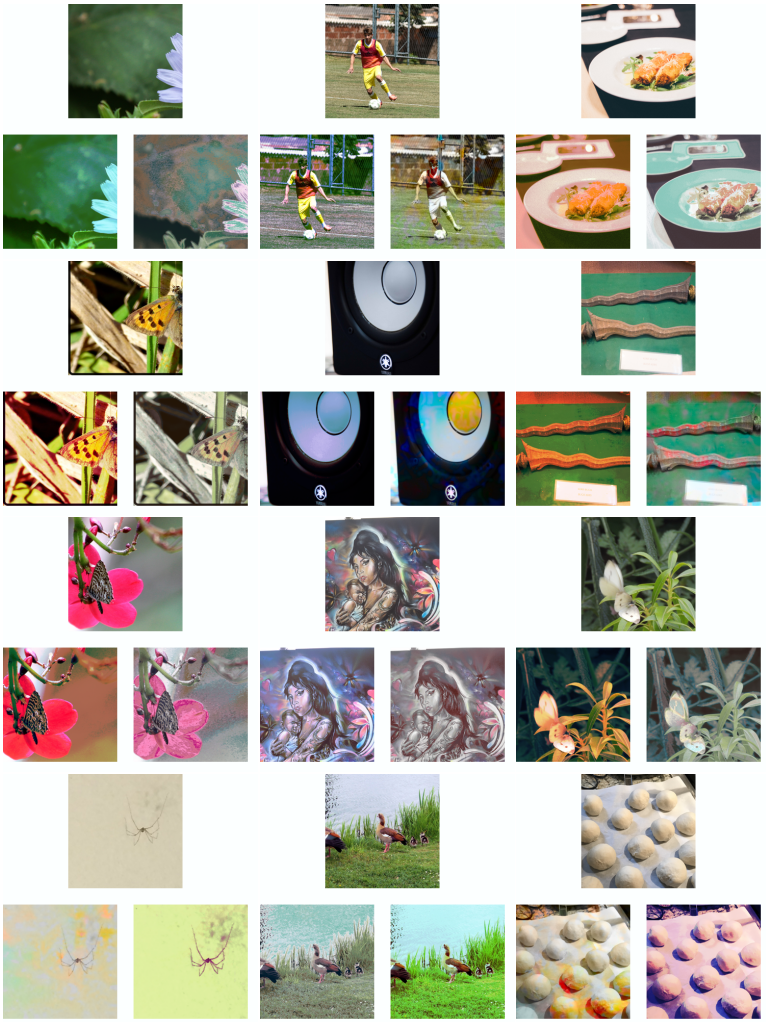}
\captionof{figure}{Examples of cases in which our AdvCF was consistently (with at least 8 of 10 workers agreed) chosen to be more visually acceptable.}
\label{fig:app_user_2}
\end{minipage}

\clearpage
\newpage

\noindent\begin{minipage}{\textwidth}
\centering
\includegraphics[width=0.95\textwidth]{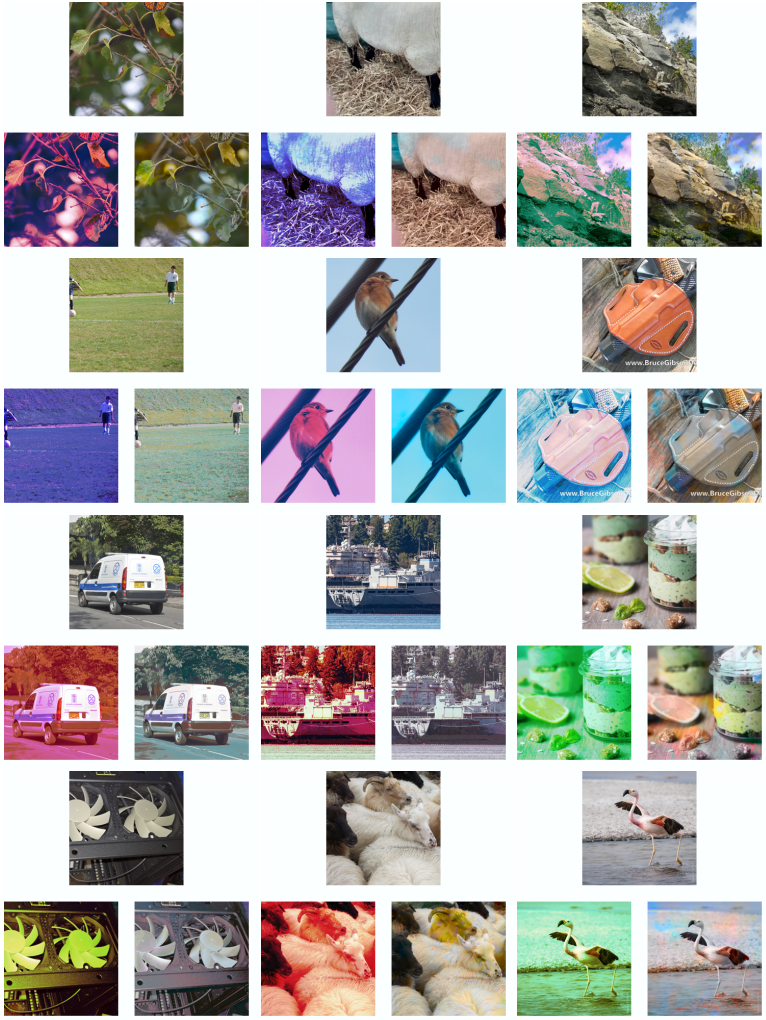}
\captionof{figure}{Examples of cases in which our AdvCF was consistently (with at least 8 of 10 workers agreed) chosen to be less visually acceptable.}
\label{fig:app_user_3}
\end{minipage}


\clearpage
\newpage

\section{Additional Experimental Results}
\label{app:add}

\begin{figure}[hbt!]
\centering
  \includegraphics[width=0.9\columnwidth]{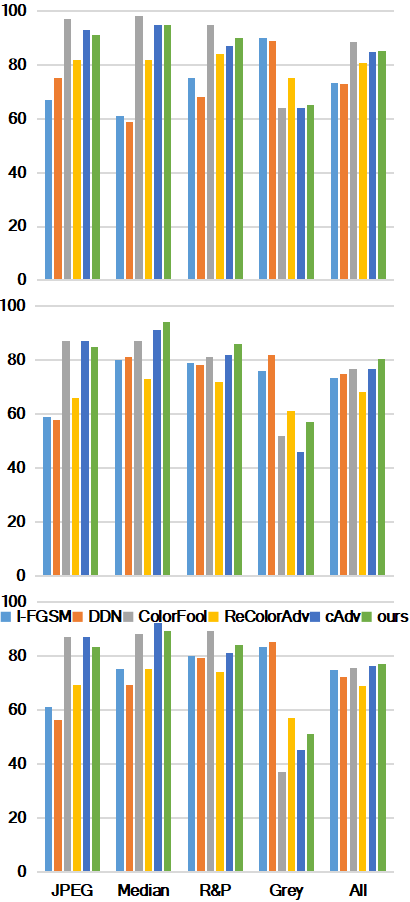}
  \caption{Proportions (\%) of adversarial images that are remained successful after input transformations: JPEG compression (Q=30)~\cite{das2018shield}, Median filtering ($3\times3$)~\cite{xu2017feature}, Resizing\&Padding~\cite{xie2017mitigating}, and Gray-scale conversion. Results are shown for three cases with different target models: AlexNet (top), VGG19 (middle), and DenseNet121 (bottom).}
\label{fig:input_trans_app}
\vspace{-0.2cm}
\end{figure}

\newpage

\section{Adversarial Training Details}
\label{app:at}



\begin{table}[hbt!]
     \caption{Adversarial training hyperparameters.}
\renewcommand{\arraystretch}{1}
\centering
\resizebox{0.5\columnwidth}{!}{
\begin{tabular}{l|c}
\toprule[1pt]
Batch\_size&128\\
Number\_of\_epochs&30\\
Optimizer&SGD\\
Learning\_rate&0.1\\
Momentum&0.9\\
Weight\_decay&$2^{-4}$\\
\bottomrule[1pt]
\end{tabular}
}

\label{tab:AT}
\end{table}


\begin{figure*}[!b]
\centering
  \includegraphics[width=0.85\textwidth]{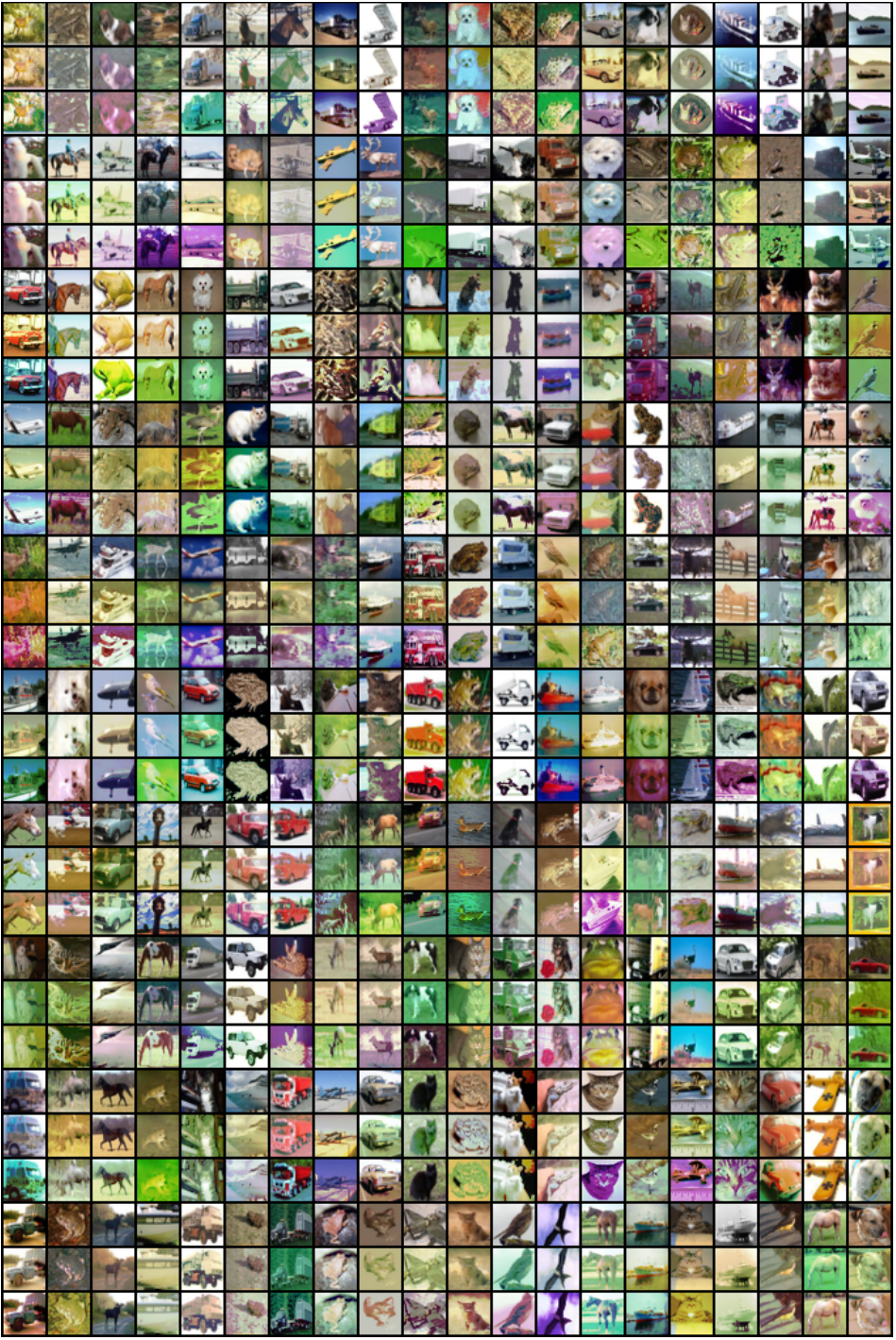}
   \caption{Adversarial images on CIFAR-10 achieved by AdvCF against undefended (second row for each example) and AdvCF-robust (third row) model. }
\label{fig:at_visual}
\vspace{-0.2cm}
\end{figure*}

\end{document}